%% file: main.tex
\documentclass[letterpaper]{article} 
\usepackage{aaai23}  
\usepackage{times}  
\usepackage{helvet} 
\usepackage{courier}  
\usepackage[hyphens]{url}  
\usepackage{graphicx} 
\usepackage{natbib} 
\usepackage{caption}
\usepackage{algorithm}
\usepackage{algorithmic}
\usepackage{multirow}
\usepackage{enumitem, array}
\usepackage{newfloat}
\usepackage{listings}
\usepackage{tikz}
\usepackage{comment}
\usepackage{amsmath,amssymb}
\usepackage{color}
\usepackage{cite}
\usepackage{subcaption}
\usepackage{booktabs}
\usepackage[utf8]{inputenc}
\usepackage{url}            
\usepackage{amsfonts}       
\usepackage{nicefrac}
\usepackage{microtype}    
\def\X{{\mathcal{X}}}
\def\Y{{\mathcal{Y}}}

\def\cP{{\mathcal{P}}}

\def\*#1{\mathbf{#1}}
\newcommand{\etal}{\textit{et al.}}
\DeclareMathOperator*{\argmax}{arg\,max}
\DeclareMathOperator*{\argmin}{arg\,min}
\DeclareCaptionStyle{ruled}{labelfont=normalfont,labelsep=colon,strut=off} 
\floatstyle{ruled}
\newfloat{listing}{tb}{lst}{}
\floatname{listing}{Listing}
\title{Distributionally Robust Optimization with Probabilistic Group}
\author{
    Soumya Suvra Ghosal, \textsuperscript{\rm 1}
    Yixuan Li \textsuperscript{\rm 1}
}
\affiliations{
    \textsuperscript{\rm 1}Department of Computer Sciences \\University of Wisconsin -- Madison\\
    \{sghosal, sharonli\}@cs.wisc.edu
   
}

\begin{document}

\maketitle

\begin{abstract}
Modern machine learning models may be susceptible to learning spurious correlations that hold on average but not for the atypical group of samples. To address the problem, previous approaches minimize the empirical worst-group risk. Despite the promise, they often assume that each sample belongs to \emph{one and only one group}, which does not allow expressing
the uncertainty in group labeling. In this paper, we propose a novel framework \textbf{PG-DRO}, which explores the idea of probabilistic group membership for distributionally robust optimization. Key to our framework, we consider soft group membership instead of hard group annotations. Our framework accommodates samples with group membership ambiguity, offering stronger flexibility and generality than the prior art. We comprehensively evaluate PG-DRO on both image classification and natural language processing benchmarks, establishing superior performance. Code is available at \url{https://github.com/deeplearning-wisc/PG-DRO}.
\end{abstract}

\input{section/introduction}

\input{section/problem_setup}

\input{section/proposed_approach}
\input{section/experiments}

\input{section/ablations}

\input{section/motivation}
\input{section/related_works}
\input{section/conclusion}
\clearpage
\input{section/acknowledge}
\bibliography{egbib}
\clearpage

\appendix
\input{section/CLIP}
\section{Training Details}
\label{app:hyperparameter}
In this section, we provide additional details on  hyperparameter tuning for both pseudo labeling and robust training.
\paragraph{Pseudo labeling using supervised learning.} 
The configurations used for pseudo group labeling are summarized in Table~\ref{tab:pseudo_hyper}. For CivilComments-WILDS dataset, we capped the number of tokens per example at 300.
\input{tables/pseudo_hyper}

\paragraph{Robust training.} During robust training with PG-DRO, we perform the following hyperparameter sweeps:

\begin{itemize}
    \item \textbf{Vision Datasets.} For both image classification tasks, Waterbirds~\cite{sagawa2019distributionally} and CelebA~\cite{liu2015deep}, we search the learning rate ($lr$) from \{$10^{-3}$, $10^{-4}$, $10^{-5}$\} and $l_2$ penalty from \{1, 0.1, $10^{-4}$\}. The generalization adjustment parameter $C$ is tuned over $\{0, 1, 2, 3, 4, 5\}$ for both datasets.
    
    \item \textbf{NLP Datasets.} For MultiNLI~\cite{williamsnli}, we directly use the hyperparameter values reported in \cite{sagawa2019distributionally}. For CivilComments-WILDS~\cite{borkancivil, koh2021wilds} we use the same configuration as used in the pseudo labeling phase except using batch size 16. Similar to vision datasets, we tune $C$ over $\{0, 1, 2, 3, 4, 5\}$ for both datasets.
\end{itemize}
We tabulate the best configurations for robust training in Table~\ref{tab:robust_hyper}.
\input{tables/robust_hyper}

\section{Ablations on Adjustment Parameter}
\label{app:adjustment_ablation}

\begin{figure*}[h!]

\begin{subfigure}{0.5\linewidth}
  \centering
  \includegraphics[width=0.75\linewidth]{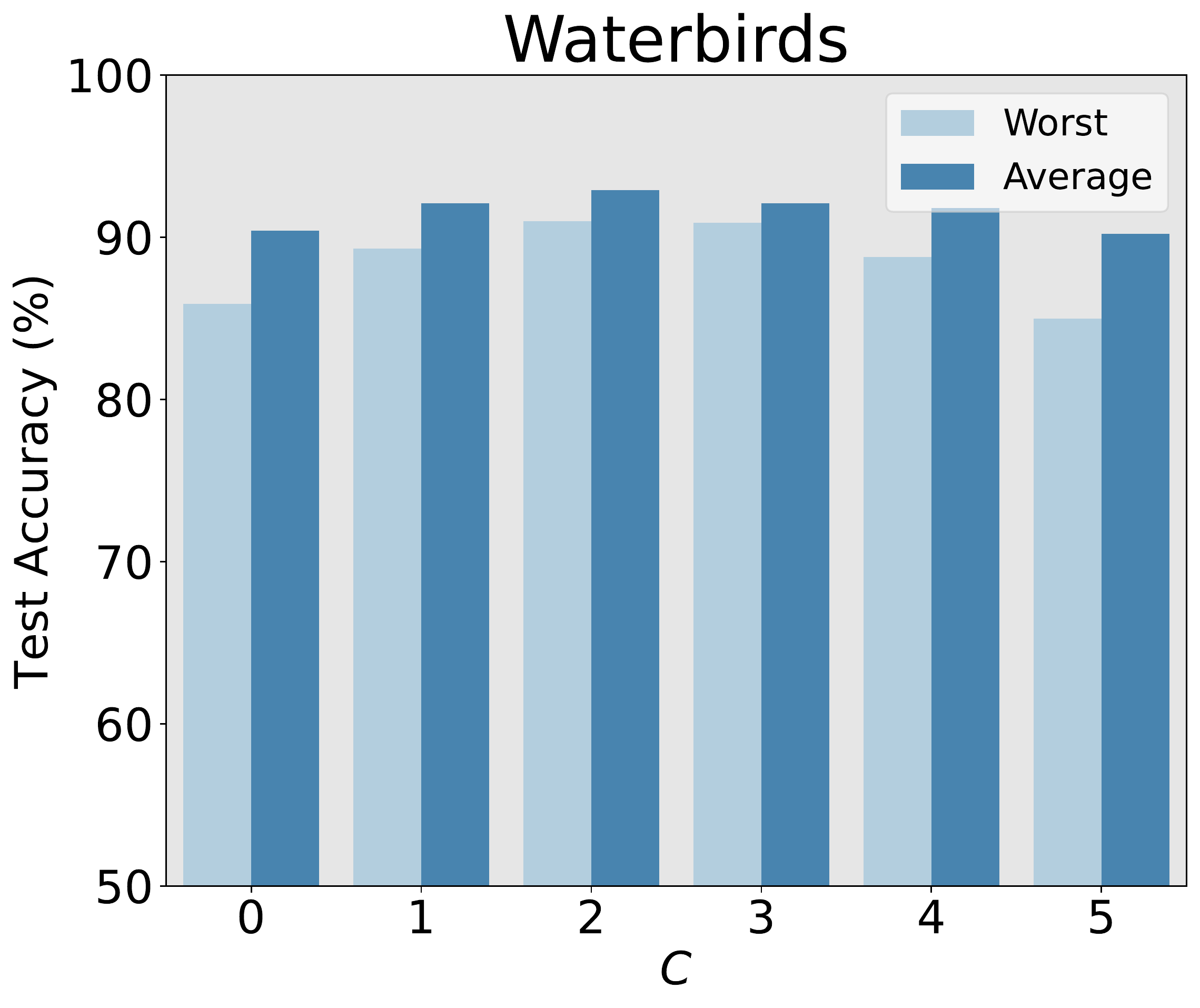}
  \label{fig:waterbirds}
\end{subfigure}%
\begin{subfigure}{0.5\linewidth}
    \centering
  \includegraphics[width=0.75\linewidth]{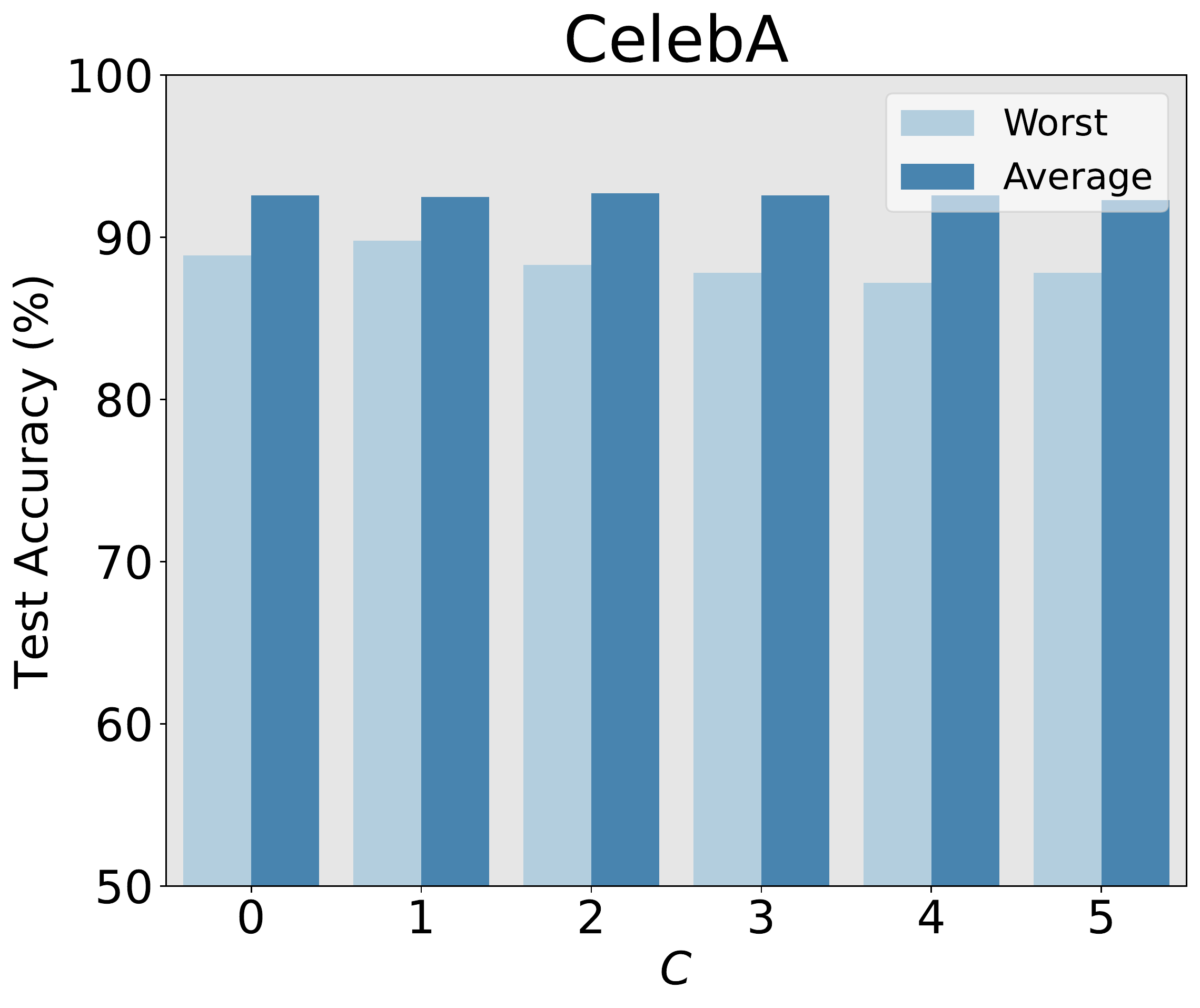}
  \label{fig:celeba}
\end{subfigure}
\caption{Ablations on the generalization adjustment parameter $C$ for Waterbirds~\cite{sagawa2019distributionally} (left) and CelebA~\cite{liu2015deep} (right). We observe that using an tuned value of adjustment parameter $C$ provides boost to the worst-group accuracy for both datasets.} 
\label{fig:ablations_c}
\end{figure*}

In this section, we perform experiments to ablate on  generalization adjustment parameter $C$ in Equation~\ref{eq:risk}. Specifically, in Figure~\ref{fig:ablations_c}, we report average and worst-group test accuracies  on Waterbirds~\cite{sagawa2019distributionally} (left) and CelebA~\cite{liu2015deep} (right) for different values of $C$. 

We observe that for the Waterbirds dataset, 
using the generalization adjustment term provides a significant improvement in worst group test accuracy. Specifically, we see a {5.1}\% increase in worst group accuracy using $C=2$ as compared to not using the adjustment parameter ($C = 0$). Setting a high value of $C$ may result in under-fitting the majority groups, thereby negatively impacting worst-group accuracy.

\section{Comparing Different Pseudo-labeling Techniques}
\label{app:pseudo}

In this section, we provide a brief comparison between two pseudo labeling approaches, based on: (1) Supervised learning~\cite{nam2022spread}, and (2) Zero-shot group classification using the large-scale pre-trained CLIP~\cite{radford2021learning} model. 

In Table~\ref{tab:spurious_pred_waterbirds}, we report group-wise accuracies for each pseudo labeling technique on Waterbirds~\cite{sagawa2019distributionally}. For evaluation, we compare the predicted group (with maximum confidence) vs. the hard group annotation. 
 
On CelebA~\cite{liu2015deep}, we observe that group-wise accuracy  using CLIP is on par with the semi-supervised learning approach despite being completely group-annotation-free. Considering Waterbirds~\cite{sagawa2019distributionally}, we see that the CLIP model performs relatively poor for the under-represented group (\textsc{waterbird} on \textsc{land}). This can be attributed to the fact that ambiguity in the group membership for these samples is difficult to capture using simple text prompts, without using much prior information. However, using probabilistic group membership in such cases helps alleviate the problem.

Further, in Table~\ref{tab:runtime} we tabulate the time required for both pseudo labeling techniques. For this benchmark, we run all experiments on a machine with 16 CPU cores and a single Nvidia Quadro RTX 5000 GPU. We can observe that being completely zero-shot, pseudo labeling using CLIP is much faster as compared to training using semi-supervised approaches.
\input{tables/spurious_predictor}

\input{tables/runtime}

\section{Additional Results}
\label{app:additional}

\subsection{GradCAM Visualizations}
\label{app:gradcam_vis}
\begin{figure*}
    \centering
    \includegraphics[width = 0.6\linewidth]{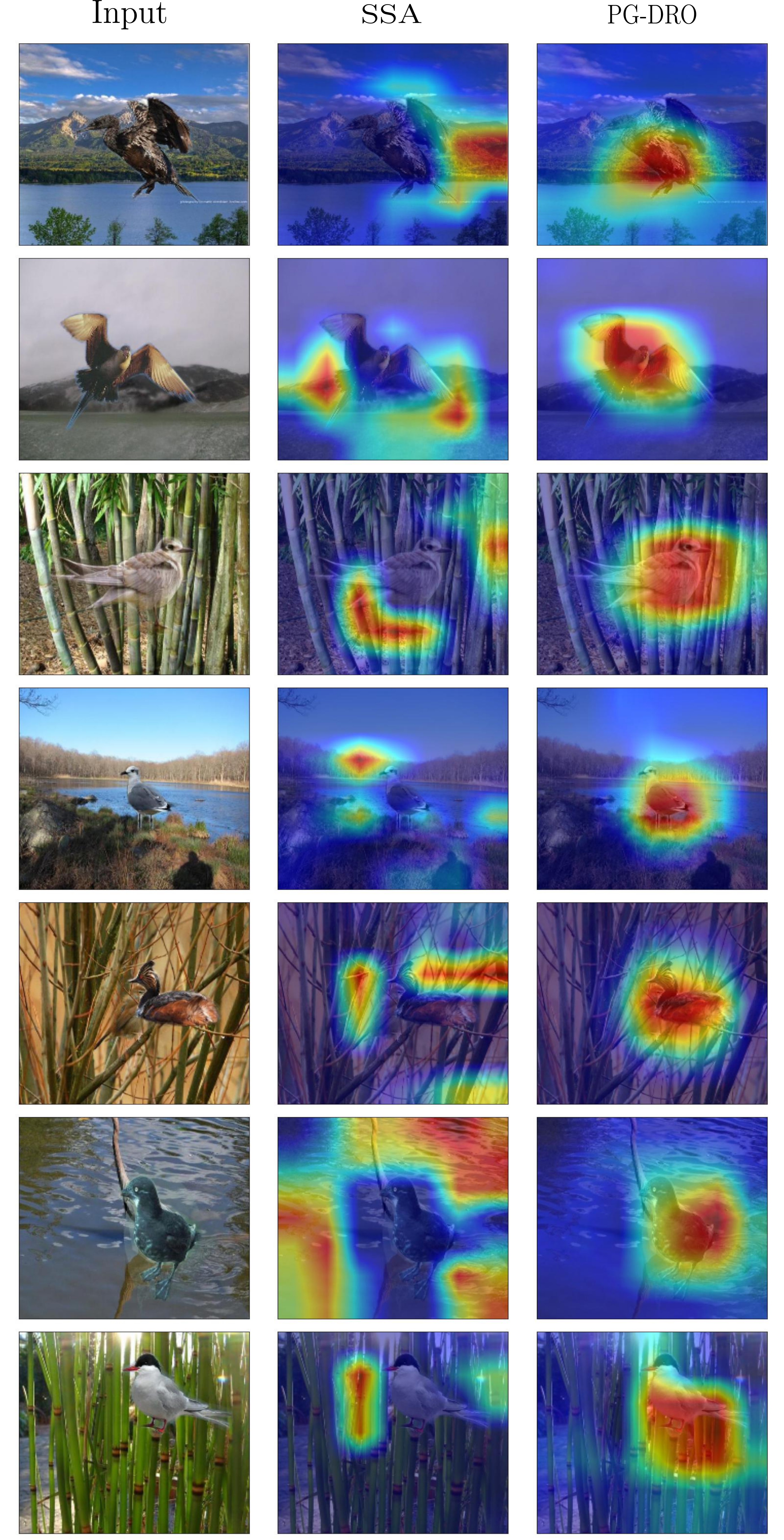}
    \caption{GradCAM visualizations for few images sampled from Waterbirds dataset~\cite{sagawa2019distributionally}. We use GradCAM~\cite{gradcam} to visualize the “salient” observed features used to classify images.}
    \label{fig:gradcam}
\end{figure*}

In Figure~\ref{fig:gradcam}, we show GradCAM~\cite{gradcam} visualizations of few representative examples from Waterbirds~\cite{sagawa2019distributionally} dataset. Specifically, we compare models trained using SSA~\cite{nam2022spread} and our proposed approach PG-DRO. For each input image, we show saliency maps where warmer colors depict higher saliency. We observe that PG-DRO consistently focuses on semantic regions representing essential cues for accurately identifying the foreground object such as claw, wing, beak, and fur. In contrast, the model trained using SSA~\cite{nam2022spread} tends to output higher salience for spurious background attribute pixels. Given that both SSA~\cite{nam2022spread} and PG-DRO uses a similar pseudo-labeling technique, this comparison highlights the significance of using group probabilities rather than one-hot labeling,

\subsection{Visualization of Learned Representations}
\begin{figure*}
    \centering
    \includegraphics[width = 0.6\linewidth]{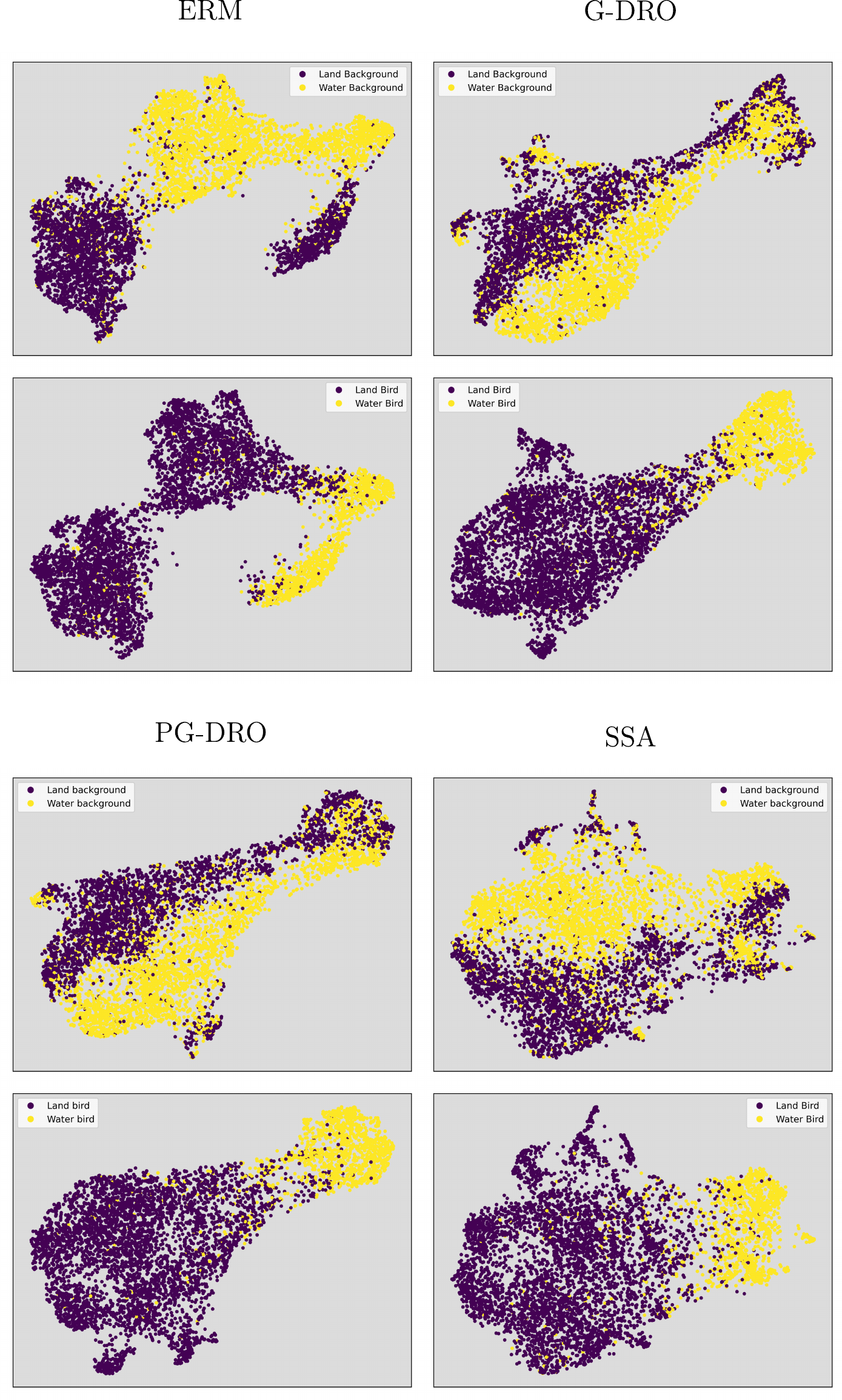}
    \caption{UMAP visualizations of learned representations for Waterbirds~\cite{sagawa2019distributionally} dataset. Specifically, we compare ERM, G-DRO~\cite{sagawa2019distributionally} and SSA~\cite{nam2022spread} with our proposed approach PG-DRO. We color each data point based on (1) spurious environment ($\mathcal{E}$) (top) and (2) true class label ($\mathcal{Y}$)  (bottom). }
    \label{fig:learned_repr}
\end{figure*}

In Figure~\ref{fig:learned_repr}, we visualize the learned representations of test samples for models trained using ERM, SSA~\cite{nam2022spread}, G-DRO~\cite{sagawa2019distributionally} and our proposed framework PG-DRO. For each method, we show two UMAP plots. Specifically, we color each data point based on (1) spurious environment ($\mathcal{E}$) (top) and (2) true class label ($\mathcal{Y}$)  (bottom). 
We observe that the model trained using ERM is highly dependent on environmental features for its prediction, signifying low worst group accuracy. Compared to ERM, both SSA and PG-DRO learns representations with more separability by $\mathcal{Y}$ as compared to $\mathcal{E}$.

\section{Software and Hardware}
\label{app:software}
We run all experiments with Python 3.7.4 and PyTorch 1.9.0. For Waterbirds~\cite{sagawa2019distributionally} and CivilComments~\cite{borkancivil, koh2021wilds}, we use Nvidia Quadro RTX 5000 GPU. Experiments on CelebA~\cite{liu2015deep} and MultiNLI~\cite{williamsnli} are run on a Google cloud instance with 12 CPUs and one Nvidia Tesla A100 GPU. 

\input{tables/analysis_vision}

\begin{figure*}
    \centering
    \includegraphics[width = 0.8\linewidth]{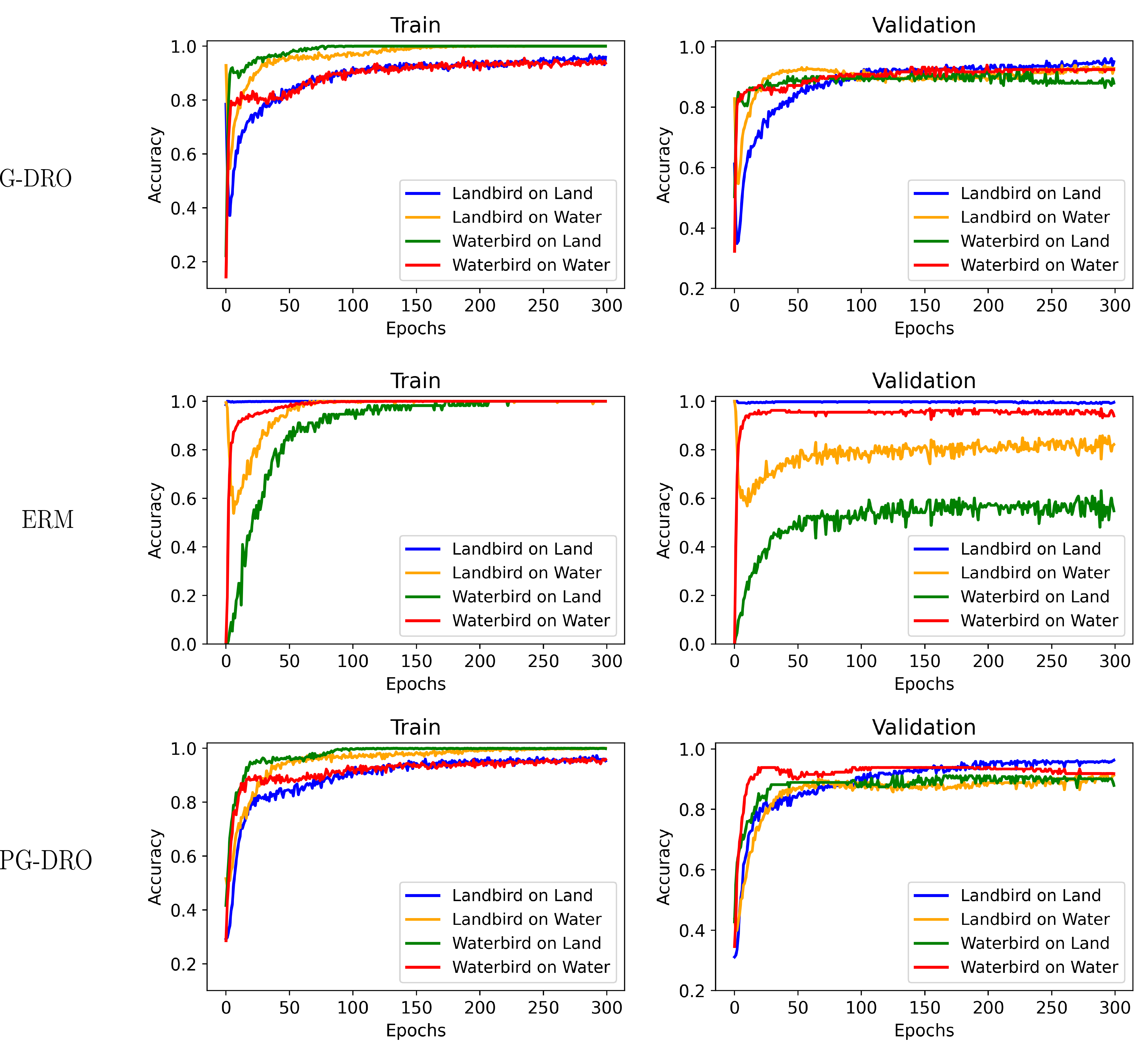}
    \caption{Training and validation accuracies for ERM, G-DRO and PG-DRO on Waterbirds~\cite{sagawa2019distributionally} dataset. As expected ERM trained model achieves perfect training accuracy across groups, but fails to generalize on worst-case group (\textsc{waterbird} on \textsc{land}).}
    \label{fig:loss}
\end{figure*}

\end{document}

%% file: section/introduction.tex
\section{Introduction}
\label{sec:intro}
A major challenge in training robust models is the presence of \emph{spurious correlations}---misleading heuristics imbibed within the training dataset
that are correlated with most examples but do not hold in general. For example, consider the Waterbirds dataset~\cite{sagawa2019distributionally}, which involves classifying bird images as {waterbird} or {landbird}. Here, the target label (bird type) is spuriously correlated with the background, \emph{e.g.}, an image of \textsc{waterbird} has a higher probability to appear on \textsc{water} background. Machine learning models, when trained on such biased datasets using empirical risk minimization (ERM), can achieve high average test accuracy but fail significantly on rare and untypical test examples lacking those heuristics (such as \textsc{waterbird} on \textsc{land})~\cite{sagawa2019distributionally, geirhos2018imagenet, sohoni2020no}. Such disparities in model prediction can lead to serious ramifications in applications where fairness or safety are important, such as facial recognition~\cite{buolamwini2018gender} and medical imaging~\cite{oakden2020hidden}. This calls for the need of ensuring {group robustness}, \emph{i.e.}, high accuracy on the under-represented groups.

\begin{figure}
  \begin{center}
    \includegraphics[width=0.32\textwidth]{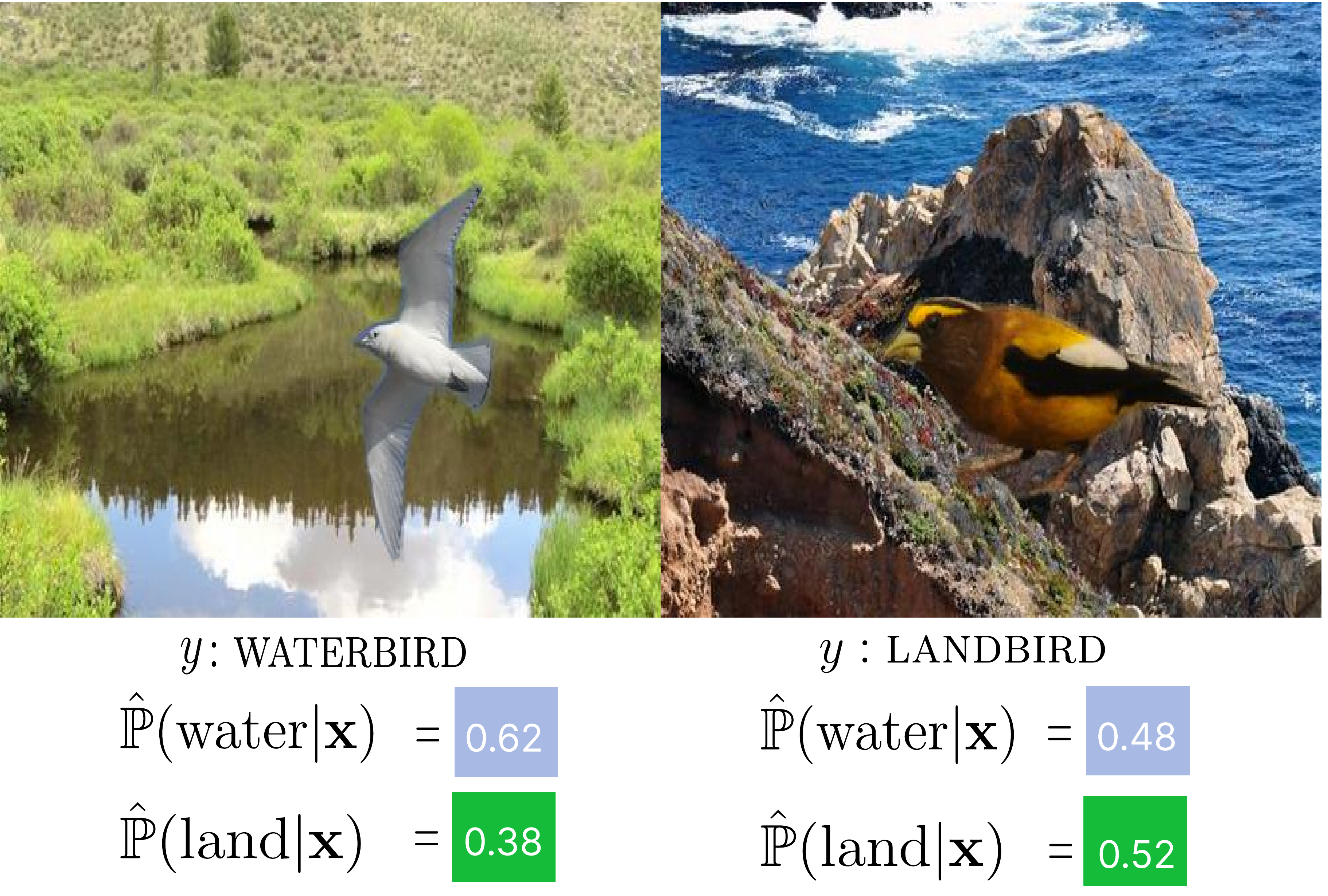}
  \end{center}
  \caption{\small Visual illustration group membership ambiguity. Images are from Waterbirds~\cite{sagawa2019distributionally} dataset.}
  \label{fig:proposed_approach_visual}
\end{figure}

Over the past few years, a  line of algorithms~\cite{sagawa2019distributionally, zhang2020coping, mohri2019agnostic, goel2020model} have been proposed to improve group robustness. The core idea behind one of the most common algorithms, G-DRO~\cite{sagawa2019distributionally}, involves minimizing the loss for the worst-performing group during training. Despite the promise, existing approaches suffer from a fundamental limitation --- they assume each sample belongs to \emph{one and only one group}, which does not allow expressing the uncertainty in group labeling. For example, in Figure~\ref{fig:proposed_approach_visual}, we show samples from the Waterbirds dataset~\cite{sagawa2019distributionally}, where the background consists of features of \emph{both} {land} and {water}, displaying inevitable ambiguity. In such cases of group ambiguity, using hard group labels can result in the loss of information in robust optimization, and disproportionally penalize the model. Taking the \textsc{landbird} in Figure~\ref{fig:proposed_approach_visual} (right) as an example, a hard assignment of the water attribute would incur an undesirably high sample-wise loss for being associated with the land attribute. 
To date, few efforts have been made to resolve this.

Motivated by this, we propose a novel framework \textbf{PG-DRO}, which performs a distributionally robust optimization using \emph{probabilistic groups}. Our framework emphasizes the uncertain nature of group membership, while minimizing the worst-group risk.  
Our key idea is to introduce the ``soft'' group membership, which relaxes the hard group membership used in previous works. 
The probabilistic group membership can allow input to be associated with multiple groups, instead of always selecting one group during robust optimization. We formalize the idea as a new robust optimization objective PG-DRO, which scales the loss for each sample based on the probability of a sample belonging to each group. Our formulation thus accommodates samples with group membership ambiguity, offering stronger flexibility and generality than G-DRO~\cite{sagawa2019distributionally}.

As an integral part of our framework, PG-DRO tackles the challenge of estimating the group probability distribution. We aim to provide a solution with (almost) full autonomy without expensive manual group annotation. Our proposed training algorithm PG-DRO can be flexibly used in conjunction with a variety of pseudo labeling approaches generating group probabilities. Our method henceforth alleviates the heavy data requirement in G-DRO, where each sample needs to be annotated with a group label. In particular, we showcase multiple instantiations through pseudo labeling approach using a small amount of group-labeled data (Section~\ref{sec:ablations}) or even without any group-annotated data (Appendix~\ref{sec:clip}). In all cases, PG-DRO achieves performance comparable to or outperforms  G-DRO~\cite{sagawa2019distributionally}, while significantly reducing group annotation.

Extensive experiments support the superiority of using group probabilities to minimize the worst-group loss. We comprehensively evaluate PG-DRO on both image classification (Section~\ref{sec:cv}) and natural language processing (Section~\ref{sec:nlp}) benchmarks, where PG-DRO establishes \emph{state-of-the-art} performance. On CelebA~\cite{liu2015deep}, using only 5\% of validation data ($988$ samples), PG-DRO achieves worst-group accuracy of {89.4}\% and outperforms G-DRO ({88.7}\%) requiring group-annotation on entire train and validation set ($182637$ samples).
Lastly, we provide a better understanding of the benefits of our method (Section~\ref{sec:simulation}), and validate that PG-DRO effectively learns a more robust decision boundary than G-DRO. 

Our \textbf{key contributions} are summarized as follows:
\begin{enumerate}
    \item We propose {PG-DRO}, a novel distributionally robust optimization framework for enhancing group robustness. 
    During training, PG-DRO leverages probabilistic group membership, which allows expressing the ambiguity and uncertainty in group labeling. To the best of our knowledge, we are the first to explore the idea of assigning group probabilities as opposed to hard group membership.
    
    \item We perform extensive experimentation on a set of vision and language processing datasets to understand the efficacy of PG-DRO. Specifically, we compare PG-DRO against the five best-performing learning algorithms to date. For both computer vision and NLP tasks, PG-DRO consistently outperforms competing methods and establishes superior performance.
    
    \item We provide extensive ablations to understand the impact of each component in our proposed framework: pseudo group labeling and robust optimization. We observe that irrespective of the pseudo group labeling approach used, PG-DRO consistently outperforms G-DRO in all cases.

\end{enumerate}

%% file: section/problem_setup.tex
\section{Preliminaries}
\label{sec:prelims}
In this paper, we consider the problem of learning a classifier when the training data has  correlations between true labels and spurious attributes. More generally, spurious attributes refer to statistically informative features that work for the majority of training examples but do not necessarily capture cues related to the labels~\cite{sagawa2019distributionally, geirhos2018imagenet,goel2020model, lifuempirical}. Recall the setup in \textsc{waterbird} vs \textsc{landbird} classification problem, where the majority of the training samples has target label (\textsc{waterbird} or \textsc{landbird}) spuriously correlated with the background features (\textsc{water} or \textsc{land} background). Sagawa \textit{et al.}~\cite{sagawa2019distributionally} showed that deep neural networks can rely on these spurious features to achieve high accuracy on average, but fail significantly for groups where such correlations do not hold.

\vspace{0.2cm}
\noindent\textbf{Problem Setup.} Formally, we consider a training set $\mathcal{D}_\text{train}$ consisting of $N$ training samples: $\{\*x_i, y_i\}^N_{i=1}$. The  samples are drawn \emph{i.i.d.} from a probability distribution: $\cP_{X,Y}$. Here, $X\in \X$ is a random variable defined in the input space, and $Y\in\Y = \{1,\ldots,K\}$ represents its label. We further assume that the data is sampled from a set of $E$ environments  $\mathcal{E} = \{e_1, e_2, \cdots, e_E\}$. The training data has spurious correlations, if the input $\*x_i$ is generated by a combination of invariant features $\*z^\text{inv}_i \in \mathbb{R}^{d_\text{inv}}$, which provides essential cues for accurate classification, and environmental features $\*z^{e}_i \in \mathbb{R}^{d_e}$ dependent on environment $e$: 
\begin{align*}
    \*x_i = \rho(\*z^\text{inv}_i, \*z_i^e).
\end{align*}
Here $\rho$ represents a function transformation from the feature space $[\*z^\text{inv}_i, \*z_i^e]^\top$ to the input space $\X$. 
Under the data model, we form groups $g = (y,e) \in  \mathcal{Y}\times  \mathcal{E} =: \mathcal{G}$ that are jointly determined by the label $y$ and environment $e$. 

Based on the data model, each sample $(\*x,y,g)$ can be denoted as a tuple consisting of input $\*x \in \X$, label $y \in \Y$, and group label $g \in \mathcal{G}$. The standard aim is to train a parameterized model $f_{\theta}: \X \rightarrow \Y$ that minimizes the expected loss $\mathbb{E}_{(\*x,y,g) \sim \cP}[l(f_{\theta}(\*x),y)]$ under the training distribution $\cP$, for some loss function $l: \Y \times \Y \rightarrow \mathbb{R}_{+}$. 

\paragraph{G-DRO.} \citet{sagawa2019distributionally} proposed group distributionally robust optimization (G-DRO), which minimizes the maximum of the expected loss among the groups:
\begin{align*}
    \mathcal{R}_\text{G-DRO}(\theta) = \max_{g \in \mathcal{G}} \mathbb{E}_{(\*x,y) \sim \cP_g}[l(f_{\theta}(\*x),y)],
\end{align*}
where $\cP_g$ indexed by $g \in \mathcal{G}=\{1,2,\cdots,|\mathcal{G}|\}$ denotes group-conditioned data distribution. Given a dataset consisting of $N$ training points $\{\*x_i, y_i, g_i\}^N_{i=1}$, G-DRO minimizes the empirical worst-group risk:
\begin{align*}
      \hat{f}_{\text{G-DRO}} =  \argmin_{\theta \in \Theta} \{\max_{g \in \mathcal{G}} \cfrac{1}{n_g} \sum_{i=1}^{N} \mathbb{I}\{g_i = g\} l(f_{\theta}(\*x_i),y_i)\},
\end{align*}
where $n_g$ represents the number of samples in each group $g \in \mathcal{G}$. In particular, G-DRO assumes that a given sample $\*x$ can belong to \emph{only one group} $g$, which does not allow expressing the
uncertainty in group labeling. 
To see this, in Figure~\ref{fig:proposed_approach_visual}, we show examples from Waterbirds dataset~\cite{sagawa2019distributionally}, where the background is associated with both \textsc{land} and \textsc{water}. Using hard group labels in such cases, can encode imprecise information about environmental attributes.

%% file: section/proposed_approach.tex
\section{Proposed Method}
\label{sec:proposed_approach}

We propose a novel distributional robust optimization framework with probabilistic group (dubbed \textbf{PG-DRO}). Our framework emphasizes the probabilistic nature of group membership, while minimizing the worst-group risk. We proceed to describe the method in detail. 

\subsection{Robust Optimization Objective}
\paragraph{Probabilistic Group.} We first introduce the notion of \emph{probabilistic group membership}, which relaxes the hard group annotation used in G-DRO. Our key idea is to consider the ``soft'' probabilities for an input $\*x$ to be in environments $\mathcal{E} = \{e_1, e_2, \cdots, e_E\}$. Formally, we denote $\hat{\mathbb{P}}(e_i|\*x)$ as the \emph{estimated} probability of an input $\*x$ associated with environment $e_i$, where:
\begin{equation}
    \sum_{i=1}^E \hat{\mathbb{P}}(e_i|\*x) = 1.
\end{equation}
Here we temporarily assume we have some estimation of the probability and we will describe the means of estimation in the following Section~\ref{sec:pseudo-group}.

Given $\hat{\mathbb{P}}(e_i|\*x)$, the probabilistic group label is defined as a vector $\hat{\mathbb{Q}}(\*x) \in \mathbb{R}^{|\mathcal{G}|}$ such that:
\begin{align}
\label{eq:defq}
     \hat{\mathbb{Q}}(\*x)_g  =
    \begin{cases}
     \hat{\mathbb{P}}(e_i|\*x)  & \text{if $g = (y,e_i)$,} \\
     0 & \text{otherwise} 
    \end{cases}
\end{align}
    where $g=\{1,2,\cdots,|\mathcal{G}|\}$ represents group index, and $|\mathcal{G}| = |\mathcal{Y}|\times|\mathcal{E}|$ indicates total number of groups. Note that our definition generalizes the hard group labeling used in G-DRO, where  $\hat{\mathbb{Q}}(\*x)_g=1$ for the assigned group and 0 elsewhere. 

\paragraph{Probabilistic Group-DRO.} With the definition above, we are ready to introduce our new learning objective called {Probabilistic Group-DRO} (PG-DRO).
Specifically, given $N$ training examples comprising triplets $\{\*x_i, y_i, \hat{\mathbb{Q}}(\*x_i)\}^N_{i=1}$, we define the empirical worst-group risk as :
\begin{align}
\label{eq:risk}
    \mathcal{\hat{R}}_\text{PG-DRO}(\theta) = \max_{g \in \mathcal{G}} \left\{ \cfrac{1}{\widetilde{n}_g} \sum_{i=1}^{N} \hat{\mathbb{Q}}(\*x_i)_g \cdot l(f_{\theta}(\*x_i),y_i) + \cfrac{C}{\sqrt{\widetilde{n}_g}} \right\},
\end{align}
where $\hat{\mathbb{Q}}(\*x_i)_g$ represents the probability of an input $\*x_i$ belonging to group $g$, and $C$ is a hyper-parameter modulating model capacity. In effect, the group probability is used as the co-efficient to scale the sample-wise loss. For any group $g \in \mathcal{G}$, we define $\widetilde{n}_g = \sum_{i=1}^{N} \hat{\mathbb{Q}}(\*x_i)_g $  as the sum of membership of all samples to group $g$. The scaling term $1/\sqrt{\widetilde{n}_g}$ constraints the model from overfitting the larger groups and focus on smaller groups. The adjustment parameter $C/\sqrt{\widetilde{n}_g}$ helps reduce the generalization gap for each group~\cite{sagawa2019distributionally}. Finally we obtain the optimal model by minimizing the risk in Equation~\ref{eq:risk}:
\begin{align*}
    \hat{f}_{\text{PG-DRO}} = \argmin_{\theta \in \Theta} \mathcal{\hat{R}_{\text{PG-DRO}}}(\theta).
\end{align*}
Note, when $\hat{\mathbb{Q}}(\*x)$ tends to one-hot encoding for all samples in the training set, indicating that a given input $\*x$ is absolutely certain to belong to a particular group, $\mathcal{\hat{R}_{\text{PG-DRO}}}(\theta)$ reduces to $\mathcal{\hat{R}}_\text{G-DRO}(\theta)$. PG-DRO henceforth provides a more general and flexible formulation than G-DRO. 

\subsection{Pseudo Group Labeling}
\label{sec:pseudo-group}
In PG-DRO, towards estimating the probabilistic group labeling $\hat{\mathbb{Q}}(\*x)$, our aim is to propose a solution with (almost) full autonomy without expensive group annotation manually. Our method alleviates the heavy requirement in G-DRO, where each sample needs to be annotated with group information.  

 Specifically, given an input $\*x$ sampled from a set of environments $\mathcal{E} = \{e_1,e_2,...,e_E\}$, we would like to first estimate each $\hat{\mathbb{P}}(e_i|\*x)$. For this, we propose training a parameterized classifier $f_{\phi}: \X \rightarrow \mathcal{E}$, that can predict the spurious environment attribute $e$. To train the model, we use a small set of group-labeled samples, $\mathcal{D_\text{L}} = \{(
\*x_1,y_1,g_1), \cdots, (\*x_m,y_m,g_m)\}$, such that $m \ll N$.
The overall objective function consists of a supervised loss for samples in $\mathcal{D_\text{L}}$ :
\begin{align}
\label{eq:pseduo-labeling}
    \hat{f}_{\text{env}} =  \argmin_{\phi \in \Phi} \{\mathbb{E}_{(\*x,y,g)\sim{\mathcal{D}_{L}}}[\ell_{\text{CE}}(f_{\phi}(\*x), e)]\},
\end{align}
where $g = (y, e)$ and $\ell_{\text{CE}}$ represents standard cross-entropy loss. We use weighted sampling based on the frequency of the samples in $\mathcal{D_\text{L}}$. Finally, for every sample $\*x \in \mathcal{D}_{\text{train}}$, we use the trained model, $ \hat{f}_{\text{env}}$, to predict the probability estimate $\hat{\mathbb{P}}(e|\*x)$ for each environment $e \in \{e_1,e_2,...,e_E\}$. We obtain group probabilities, $\hat{\mathbb{Q}}(\*x)$, as defined in Equation~\ref{eq:defq} and use it for training PG-DRO. 

\vspace{0.1cm}
\noindent \textbf{Remark.} Note that our proposed training objective PG-DRO can be flexibly used in conjunction with other pseudo labeling approaches generating group probabilities. Beyond pseudo labeling with supervised loss above, we also showcase the strong feasibility of alternative group labeling approaches. We provide further details in Section~\ref{sec:ablations} and Appendix~\ref{sec:clip}. 

%% file: section/experiments.tex
\section{Experiments}
\label{sec:experiment}

In this section, we comprehensively evaluate PG-DRO on both computer vision tasks (Section~\ref{sec:cv}) and natural language processing (Section~\ref{sec:nlp}) containing spurious correlations. 

\subsection{Evaluation on Image Classification Benchmarks}
\label{sec:cv}

\input{tables/vision_dataset}
\input{tables/analysis_vision_2}

\noindent \textbf{Datasets.} In this study, we consider two common image classification benchmarks: Waterbirds~\cite{sagawa2019distributionally} and CelebA~\cite{liu2015deep}. 

 \textbf{(1) CelebA}~\cite{liu2015deep}: Training samples in CelebA have spurious associations between target label and demographic information such as gender. 
 We use the label space $\mathcal{Y} = \{\textsc{blond hair}, \textsc{dark hair}\}$ and gender as the spurious feature, $\mathcal{E} = \{\textsc{male}, \textsc{female}\}$. The training data consists of $162770$ images with $1387$ in the smallest group, \emph{i.e.}, \textsc{male} with \textsc{blond hair}.
  
\textbf{(2) Waterbirds}~\cite{sagawa2019distributionally}: Introduced in~\cite{sagawa2019distributionally}, this dataset contains spurious correlation between the background features and target label $y\in$ \{\textsc{waterbird}, \textsc{landbird}\}. The dataset is constructed by selecting bird photographs from the Caltech-UCSD Birds-200-2011 (CUB)~\cite{WahCUB_200_2011} dataset  and then superimposing on $\mathcal{E}=  \{\textsc{water}, \textsc{land}\}$ background selected from the Places dataset~\cite{zhou2017places}. The dataset consists of $n = 4795$ training examples, with the smallest group size 56 (\emph{i.e.}, \textsc{waterbird} on \textsc{land} background). 

\vspace{0.2cm}
\noindent \textbf{Training details.} For experimentation on image classification datasets, following prior works~\cite{sagawa2019distributionally} we use ResNet-50~\cite{He2016DeepRecognition} initialized from ImageNet pre-trained model. Models are selected by maximizing the worst-group accuracy on the validation set. \emph{We provide detailed description regarding hyper-parameter in Appendix~\ref{app:hyperparameter} and \ref{app:adjustment_ablation}}.

\vspace{0.2cm}
\noindent \textbf{Metrics.} For all methods, we report two standard metrics: (1) average test accuracy and (2) worst-group test accuracy. In particular, the worst-group test accuracy indicates the model’s generalization performance for groups where the correlation between the label $y$ and environment $e$ does not hold. High worst-group test accuracy indicates a model's less reliance on the spurious attribute.

\paragraph{PG-DRO outperforms strong baselines.} In Table~\ref{tab:vision_dataset_main}, we show that our approach PG-DRO significantly outperforms all the rivals on vision datasets, measured by the worst-group accuracy. For comparison, we include the best-performed learning algorithms to date: CVaR DRO (Levy et al. 2020), LfF (Nam et al. 2020), EIIL (Creager, Jacobsen, and Zemel
2021), JTT (Liu et al. 2021), and SSA (Nam et al. 2022)—all of which are developed without assuming the availability of group-labeled training data. Similar to ours, these methods utilize a validation set that contains the group labeling information. The comparisons are thus fair given the same amount of information. In addition, we also compare with G-DRO~\cite{sagawa2019distributionally}, which requires the entire training dataset to be labeled with group attribute and hence is significantly more expensive from an annotation perspective.

We highlight two salient observations: \textbf{(1)} On CelebA, PG-DRO outperforms G-DRO by {1.3}\% in terms of worst-group accuracy. The result signifies the advantage of using probabilistic group labeling compared to hard group labeling. Moreover, our method achieves overall better performance than G-DRO while using \emph{significantly less group annotation} (validation set only). \textbf{(2)} Among methods that do not use group-labeled training samples, PG-DRO outperforms the current SOTA methods, JTT~\cite{liu2021just} and SSA~\cite{nam2022spread}, by {4.3}\% and {2}\% on Waterbirds dataset respectively. As expected, while ERM consistently achieves the best average accuracy among all methods, its worst-group accuracy suffers the most.

\paragraph{PG-DRO remains competitive under reduced group annotation.} In Table~\ref{tab:analysis_vision_2}, we show that the benefits of using group probabilities persist even when the model is trained with reduced group annotation. Specifically, we perform controlled experiments and compare PG-DRO with two latest SOTA methods, SSA~\cite{nam2022spread} and JTT~\cite{liu2021just}, in settings where the group labels are available on $100\%$, $10\%$ and $5\%$ of the validation set. We observe that even under reduced group annotation, PG-DRO consistently outperforms both SSA and JTT. In particular, under the challenging setting with only 5\% of the validation set, PG-DRO outperforms both SSA and JTT by {2.1}\% and {13.2}\% on Waterbirds~\cite{sagawa2019distributionally} dataset, in terms of worst-group accuracy.

Further, we also include additional qualitative evidence via GradCAM~\cite{gradcam} visualizations in \emph{Appendix~\ref{app:additional}}. We observe that for the Waterbirds dataset, PG-DRO consistently focuses on semantic regions representing essential cues for accurately identifying the foreground object such as claw, wing, beak, and fur. In contrast, baseline methods tend to output higher salience for spurious background attribute pixels.

\input{tables/nlp_dataset}
\subsection{Evaluation on Natural Language Processing Tasks}
\label{sec:nlp}

\paragraph{Datasets.} To validate the effectiveness of PG-DRO, we perform experiments on two natural language processing datasets: MultiNLI and CivilComments-WILDS containing spurious correlations.

\textbf{(1) MultiNLI}~\cite{williamsnli}: The Multi-Genre Natural Language Inference (MultiNLI) dataset is a crowdsourced collection of sentence pairs with the premise and hypothesis. The label indicates whether the hypothesis is entailed by, contradicts, or is neutral to the premise. Hence, the label space is defined as $\mathcal{Y} = \{\textsc{entailed}, \textsc{neutral}, \textsc{contradictory}\}$. Previous study~\cite{gururanganannotation} has shown the presence of spurious associations between the target label \textsc{contradictory} and set of negation words, $\mathcal{E}$=\{\textsc{nobody}, \textsc{no}, \textsc{never}, \textsc{nothing}\}.

\textbf{(2) CivilComments-WILDS}~\cite{borkancivil,koh2021wilds}: For this dataset, we use a similar setup as defined in ~\cite{nam2022spread}. Each instance in this dataset corresponds to an online comment generated by users which is labeled as either toxic or not toxic, $\mathcal{Y} = \{\textsc{toxic}, \textsc{non-toxic}\}$. The spurious attribute is set as $\mathcal{E} = \{\textsc{identity}, \textsc{no identity}\}$, where \textsc{identity} indicates comment associated with any of the  demographic identities \{male, female, White, Black, LGBTQ, Muslim, Christian, other religion\}.

\paragraph{Training details.} For experimentation on language processing tasks, we use a pre-trained BERT model~\cite{devlin2018bert}. Specifically, we use the Hugging Face PyTorch-Transformers~\cite{wolf-etal-2020-transformers} implementation of the BERT \verb|bert-base-uncased| model. We use the
default tokenizer and model settings. The evaluation metrics are the same as Section~\ref{sec:cv}. \emph{Refer Appendix~\ref{app:hyperparameter} and \ref{app:adjustment_ablation} for detailed description regarding hyper-parameters.} 

\paragraph{PG-DRO achieves superior performance.} In Table~\ref{tab:nlp_dataset_main}, we provide a comprehensive comparison against an array of well-known learning algorithms designed to tackle spurious correlations in training data, including CVaR DRO~\cite{levy2020large}, LfF~\cite{nam2020learning}, EIIL~\cite{creager2021environment}, JTT~\cite{liu2021just}, and SSA~\cite{nam2022spread}. For fair evaluation, all competing methods utilize group annotations on the validation set. We also compare against standard baselines such as ERM which does not require group labeling, and G-DRO~\cite{sagawa2019distributionally} which assumes group labeling information for both train and validation set. 

Similar to our observations on vision datasets, PG-DRO upholds the trend of outperforming competing baselines achieving superior performance on both natural language processing tasks. In particular, PG-DRO improves the current best method SSA~\cite{nam2022spread} by {3.7}\% and {2.3}\% on the CivilComments-WILDS and MultiNLI dataset respectively.

%% file: tables/vision_dataset.tex
\begin{table*}[t]
\small
\centering
 \begin{tabular}{l ccccc}
\toprule
 \multirow{2}{2cm}{\textbf{Method}} &  \multirow{2}{3cm}{\centering Dataset with group label} & \multicolumn{2}{c}{\centering \textbf{Waterbirds}} & \multicolumn{2}{c}{\centering \textbf{CelebA}}\\ \cmidrule{3-6}
 & &  Avg. Acc & Worst Group Acc & Avg. Acc &  Worst Group Acc \\
 \midrule
 ERM~\cite{vapnik1991principles} & None & {97.3} & 63.2 & 95.6 & 47.2 \\ 
 CVaR DRO~\cite{levy2020large} & val. set & 96.0 & 75.9 & 82.5 & 64.4 \\
 LfF~\cite{nam2020learning} & val. set & 91.2 & 78.0 & 85.1 & 77.2 \\
 EIIL~\cite{creager2021environment} & val. set & 96.9 & 78.7 & 91.9 & 83.3 \\
 JTT~\cite{liu2021just} & val. set & 93.3 & 86.7 & 88.0 & 81.1 \\
 SSA~\cite{nam2022spread} & val. set & 92.2 & 89.0 & 92.8 & 89.8 \\
 \midrule
\textbf{PG-DRO} (Ours) & val. set & 92.5$_{\pm{0.5}}$ & \textbf{91.0}$_{\pm{0.6}}$ & 92.2$_{\pm{0.4}}$ & \textbf{90.0}$_{\pm{0.9}}$ \\
  \midrule
 G-DRO~\cite{sagawa2019distributionally} & train \& val. set & 92.4$_{\pm{0.2}}$ & 90.7$_{\pm{0.9}}$ & 92.8$_{\pm{0.3}}$ & 88.7$_{\pm{1.5}}$\\ 
 \bottomrule
 \end{tabular}
 \caption{\small Comparison of average and worst-group test accuracies for different methods when evaluated on image classification datasets: Waterbirds~\cite{sagawa2019distributionally} \& CelebA~\cite{liu2015deep}. We obtain the results of CVaR DRO, LfF, EIIL, JTT and SSA from \cite{nam2022spread}. Results (mean and std) of our method are estimated over 3 random runs. Best performing results (in terms of worst-group accuracy) are marked in \textbf{bold}.}
\label{tab:vision_dataset_main}
\end{table*}

%% file: tables/analysis_vision_2.tex
\begin{table}[t]

\small
\centering
 \begin{tabular}{l cccccc}
\toprule
 \multirow{2}{*}{\textbf{Method}} & \multicolumn{3}{c}{\centering \textbf{Waterbirds}} & \multicolumn{3}{c}{\centering \textbf{CelebA}}\\ \cmidrule{2-7}
 & $100\%$ & $10\%$ & $5\%$ & $100\%$ &  $10\%$ & $5\%$ \\
 & \multicolumn{3}{c}{\centering {(Val. set)}} & \multicolumn{3}{c}{\centering {(Val. set)}}\\
 \midrule
 JTT & 86.7 &  86.9 & 76.0 & 81.1 & 81.1 & 82.2 \\
 SSA & 89.0 &  88.9 & 87.1 & 89.8 & 90.0 & 86.7 \\
 \midrule
 PG-DRO & \textbf{91.0} & \textbf{90.3} & \textbf{89.2} & \textbf{90.0} & \textbf{90.6} & \textbf{89.4} \\ 
 \bottomrule
 \end{tabular}
 \caption{\small Worst-group accuracy on Waterbirds and CelebA under varying fractions of the group-annotated validation set. Our proposed framework, PG-DRO, outperforms state-of-art methods even under reduced group annotation. Results of JTT~\cite{liu2021just} and SSA~\cite{nam2022spread} are from \cite{nam2022spread}.}
\label{tab:analysis_vision_2}
\end{table}

%% file: tables/nlp_dataset.tex
\begin{table*}[t]
\centering
\small
\begin{tabular}{lccccc}

\toprule
 \multirow{2}{1.5cm}{\textbf{Method}} &  \multirow{2}{*}{\centering Dataset with group label } & \multicolumn{2}{c}{\centering \textbf{MultiNLI}} & \multicolumn{2}{c}{\centering \textbf{CivilComments-WILDS}}\\ \cmidrule{3-6}
 & &  Avg. Acc & Worst Group Acc & Avg. Acc &  Worst Group Acc \\
 \midrule
 ERM~\cite{vapnik1991principles} & None & 82.6 & 66.4 & 92.6 & 58.4 \\ 
 CVaR DRO~\cite{levy2020large} & val. set & 82.0 & 68.0 & 92.5 & 60.5 \\
 LfF~\cite{nam2020learning} & val. set & 80.8 & 70.2 & 92.5 & 58.8 \\
 EIIL~\cite{creager2021environment} & val. set & 79.4 & 70.9 & 90.5 & 67.0 \\
 JTT~\cite{liu2021just} & val. set & 78.6 & 72.6 & 91.1 & 69.3 \\
 SSA~\cite{nam2022spread} & val. set & 79.9 & 76.6 & 88.2 & 69.9\\
 \midrule
  \textbf{PG-DRO} (Ours) & val. set & 81.0$_{\pm{0.4}}$ & \textbf{78.9}$_{\pm{0.9}}$ & 89.2$_{\pm{1.4}}$ & \textbf{73.6}$_{\pm{1.8}}$ \\
  \midrule
 G-DRO~\cite{sagawa2019distributionally} & train \& val. set & 81.4$_{\pm{0.1}}$ & 77.5$_{\pm{1.2}}$ & 87.7$_{\pm{0.6}}$ & 69.1$_{\pm{0.8}}$ \\
 \bottomrule
 \end{tabular}
 \caption{\small Comparison of average and worst-group test accuracies for different methods when evaluated on language processing datasets: MultiNLI~\cite{williamsnli} \& CivilComments-WILDS~\cite{borkancivil,koh2021wilds}. We obtain the results of CVaR DRO, LfF, EIIL, JTT and SSA from \cite{nam2022spread}. Results (mean and std) of our method are estimated over 3 random runs. Best performing results (in terms of worst-group accuracy) are marked in \textbf{bold}.}
\label{tab:nlp_dataset_main}
\end{table*}

%% file: section/ablations.tex
\section{Further Ablations}
\label{sec:ablations}
\input{tables/ablation}
Given the strong performance of PG-DRO on both computer vision (Section~\ref{sec:cv}) and NLP (Section~\ref{sec:nlp}) tasks, we take a step further to understand the advantages of using probabilistic group membership over hard group annotations. Specifically, in this section, we design controlled experiments to ablate the role of each component in our proposed framework: pseudo group labeling and robust optimization. In particular, for comprehensive evaluation and comparison, we train the spurious attribute classifier ($\hat{f}_{\text{env}}$) using: (1) supervised approach described in Section~\ref{sec:pseudo-group}, (2) a new zero-shot approach (see details in Appendix~\ref{sec:clip}), and (3) semi-supervised learning~\cite{lee2013pseudo, nam2022spread}. The semi-supervised objective additionally requires using group unlabelled training samples, in addition to a small number of labeled ones. 

\paragraph{PG-DRO outperforms G-DRO under different pseudo group labeling methods.} In Table~\ref{tab:ablations}, we provide a comprehensive comparison between PG-DRO and G-DRO for different pseudo group labeling objectives. We observe that for both Waterbirds and CelebA datasets, PG-DRO consistently outperforms G-DRO under different approaches for pseudo group labeling. These results further highlight that the advantages of using probabilistic group membership can be leveraged irrespective of the pseudo group labeling approach used. Moreover, PG-DRO continues its dominance over G-DRO under the most challenging setting, when group annotations are available for only 5\% of the validation set. These experiments further underwrite the importance of using probabilistic group membership.

%% file: tables/ablation.tex
\begin{table*}[h]

\small
\centering
 \begin{tabular}{ccccccc}
\toprule
 \multirow{2}{*}{\shortstack{\\\textbf{Pseudo Group} \\ \textbf{Labeling Method}}} & \multirow{2}{*}{\shortstack{ \\ \textbf{Robust Optimization} \\ \textbf{Method}}} &  \multirow{2}{*}{\shortstack{\\\textbf{Dataset with group}\\ \textbf{label}}} & \multicolumn{2}{c}{\centering \textbf{Waterbirds}} & \multicolumn{2}{c}{\centering \textbf{CelebA}}\\ \cmidrule{4-7}
 & & & Avg. Acc & Worst Group Acc & Avg. Acc &  Worst Group\\
 \midrule
 \multirow{2}{*}{Supervised} & G-DRO & val. set & 92.4 & 88.0 & 91.6 & 88.6 \\ 
 & PG-DRO (Ours) & val. set & 92.5 & \textbf{91.0} & 92.2 & \textbf{90.0} \\
 \midrule
 \multirow{2}{*}{Supervised} & G-DRO & $5\%$ of val. set & 92.5 & 87.5 & 92.5 & 87.5 \\ 
 & PG-DRO (Ours) & $5\%$ of val. set & 91.7 & \textbf{89.2} & 92.0 & \textbf{89.4}\\
 \midrule
 \multirow{2}{*}{Semi-Supervised} & G-DRO & val. set & 92.2 & 89.0 & 92.8 & \textbf{89.8} \\ 
 & PG-DRO (Ours) & val. set & 92.4 & \textbf{91.0} & 92.5 & \textbf{89.8} \\
 \midrule
 \multirow{2}{*}{Semi-Supervised} & G-DRO & $5\%$ of val. set & 92.6 & 87.1 & 92.8 & 86.7 \\ 
 & PG-DRO (Ours) & $5\%$ of val. set & 91.9 & \textbf{89.2} & 91.7 & \textbf{87.8} \\
 \midrule
 \multirow{2}{*}{Zero-shot} & G-DRO & None & 91.2 & 87.6 & 93.2 & 87.8 \\ 
 & PG-DRO (Ours) & None & 91.5 & \textbf{89.4} & 92.7 & \textbf{88.8} \\
 \bottomrule
 \end{tabular}
 \caption{\small Comparison of average and worst-group test accuracies for different combinations of training objectives used during pseudo group labeling and robust optimization. Best performing results (in terms of worst-group accuracy) are marked in \textbf{bold}.}
\label{tab:ablations}
\end{table*}

%% file: section/motivation.tex
\section{Understanding Benefits of PG-DRO}
\label{sec:simulation}

The results in Section~\ref{sec:experiment} and Section~\ref{sec:ablations} validate the improved performance of our proposed framework PG-DRO. In this section, we seek to provide a better understanding for the improved performance. 
\subsection{Setup}
\label{subsec:synthetic_exp}

\paragraph{Data distribution.} We construct a synthetic dataset to better understand the advantages of using group-probabilities over hard group annotations. Compared to the complex real datasets studied in Section~\ref{sec:experiment}, this synthetic data helps us directly visualize and understand each component of PG-DRO, and its influence on the decision boundary. Specifically, the training dataset ($\mathcal{D}_{\text{train}}$) consists of $n$ training samples: $\{\*x_i, y_i\}_{i=1}^n$, where the label $y=\{-1,1\}$ is spuriously correlated with the environment attribute $e=\{-1,1\}$. Replicating the real datasets studied in Section~\ref{sec:experiment}, we divide the training data into four groups accordingly: two majority groups with $y = e$, each of size $n_{\text{maj}}/2$, and two minority groups with $y = -e$, each of size $n_{\text{min}}/2$. Further, we set $n = n_{\text{maj}} + n_{\text{min}}$ as the
total number of training points, and $p = n_{\text{maj}}/n$ as the
fraction of majority examples. A higher value of $p$ indicates a stronger correlation between target label $y$ and environment attribute $e$. 

An input sample $\*x = [z_{\text{inv}}, z_{\text{e}}] \in \mathbb{R}^{2}$ comprises of invariant features $z_{\text{inv}} \in \mathbb{R}$ generated from target label $y$, and environmental/spurious features $z_{\text{e}} \in \mathbb{R}$ generated from environment $e$. In particular, we generate:
\begin{align*}
    z_{\text{inv}} \sim \mathcal{N}(y, \sigma^2_{\text{inv}}) \\
    z_{\text{e}} \sim \mathcal{N}(e, \sigma^2_{\text{e}}), 
\end{align*}
The randomness in the invariant and environmental features are controlled through their respective variances. For all experiments in Section~\ref{subsec:obvs_synthetic_data}, we fix the total number of training points $n = 4000$ and majority fraction $p = 0.95$. Further, $\sigma^2_{\text{inv}}$ and $\sigma^2_{\text{e}}$ are set as 0.5 and 0.05 respectively to encourage the model to use the
environmental features over the invariant features.

\noindent\textbf{Model.} Both for the purpose of pseudo group labeling and robust optimization, we use a simple four layered neural network with ReLU activation. We provide further details of model training in Appendix.

\begin{figure}[h]
\centering
\begin{subfigure}{0.5\columnwidth}
  \centering
  \includegraphics[scale = 0.28]{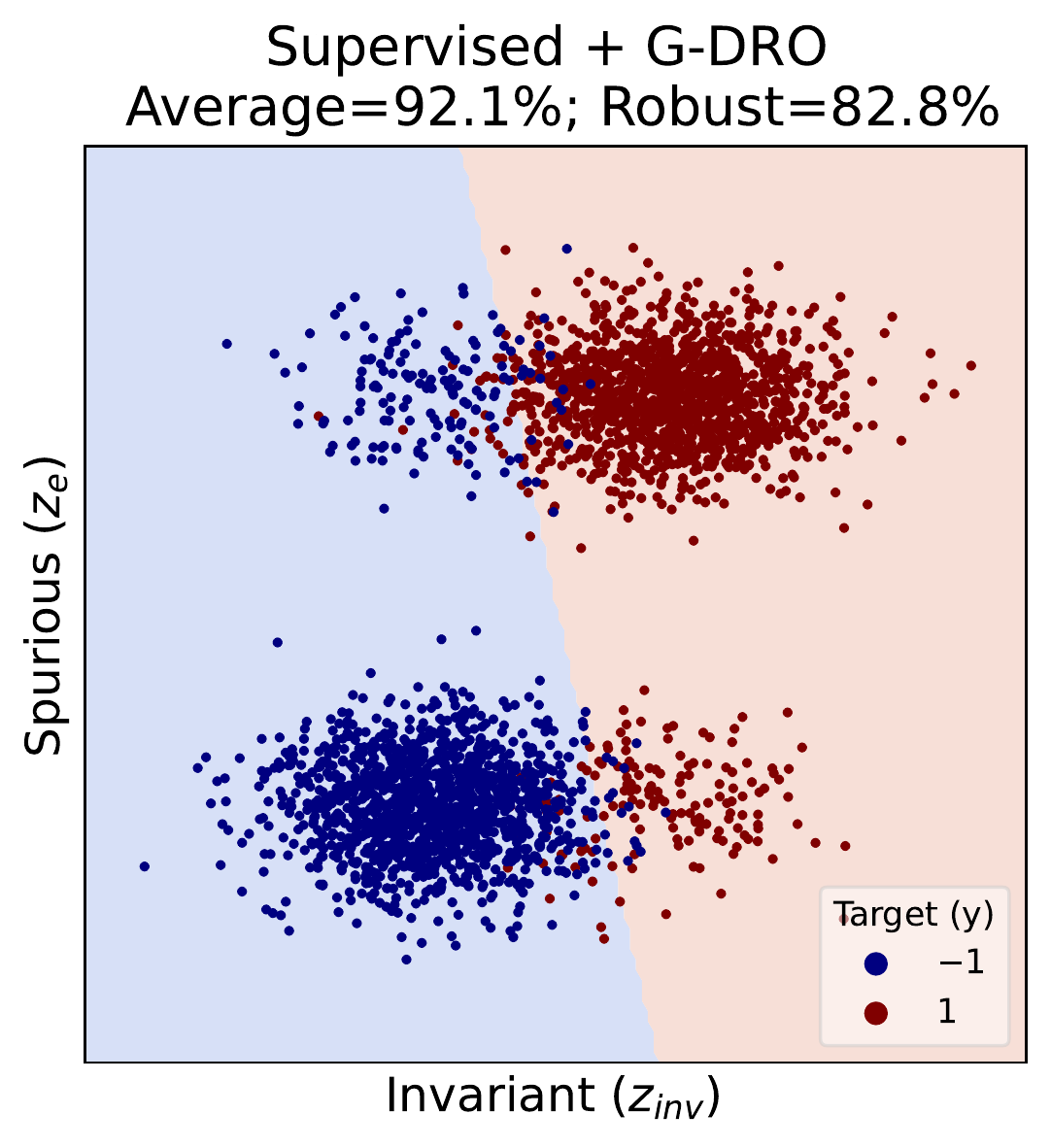}
  \caption{}
  \label{fig:gdro}
\end{subfigure}%
\begin{subfigure}{0.48\columnwidth}
 
  \includegraphics[scale = 0.28]{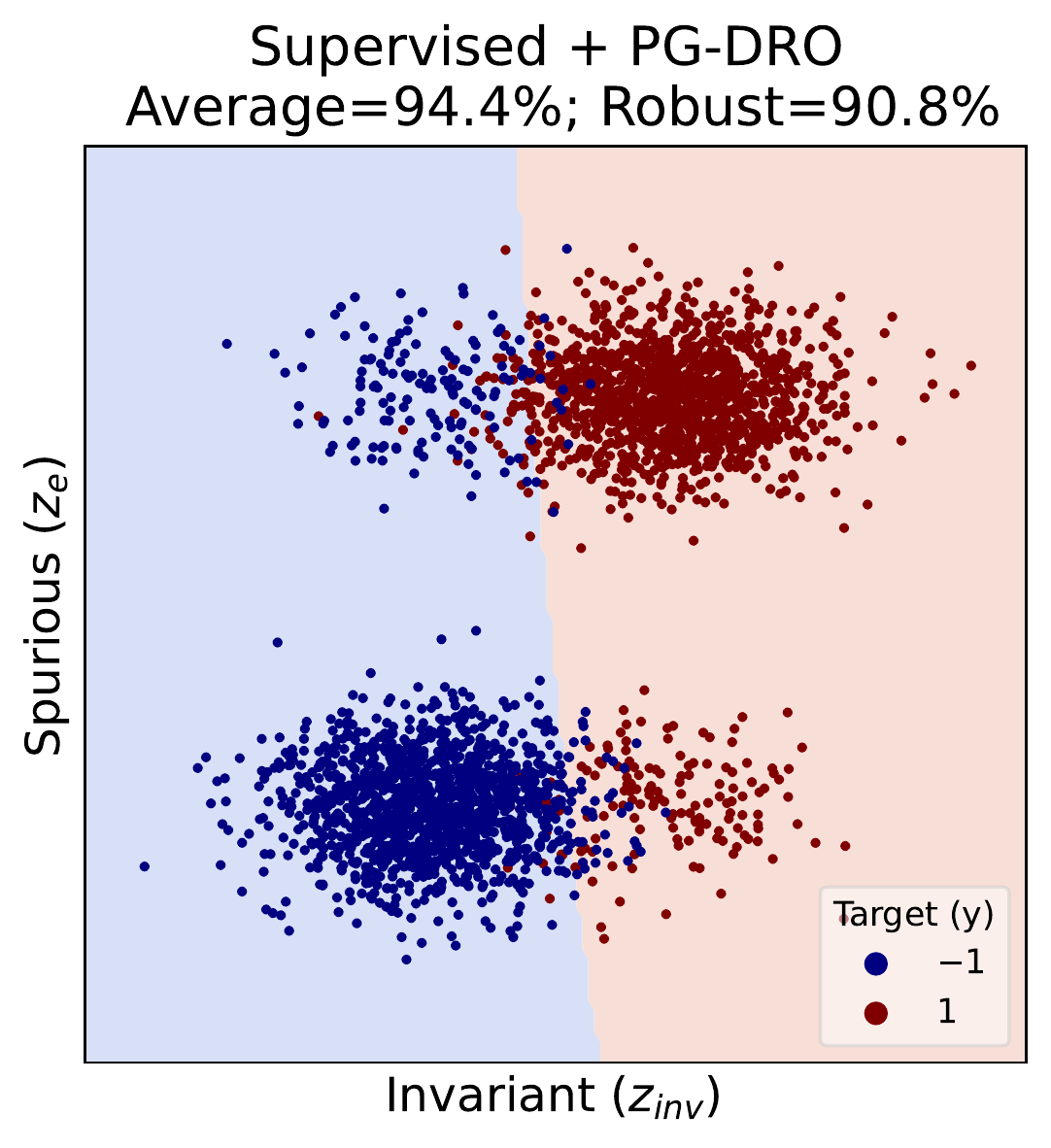}
  \caption{}
  \label{fig:pgdro}
  
\end{subfigure}
\caption{\small Consider data points $\*x$ in $\mathbb{R}^2$ with two classes $y$. The vertical axis of $\*x$ is the environment/spurious feature that highly correlates with $y$, and the horizontal axis is the invariant feature. The data consists of four groups, where the \emph{top-left and lower-right are two minority groups}. The robust accuracy is test worst-group accuracy. Model trained using PG-DRO (b) is more robust as compared to the G-DRO (a). {Color of points encodes ground truth label $y$}. Color of background shade indicates model's prediction $\hat y$.}
\label{fig:toy_img}
\end{figure}

\subsection{Insights}
\label{subsec:obvs_synthetic_data}

 \noindent \textbf{PG-DRO leads to a more robust decision boundary than G-DRO.} In this study, we train a group predictor on a small group-labeled dataset ($\mathcal{D}_{\text{L}}$) consisting of $100$ samples. 
 
 Next, the trained group classifier is used to generate both hard group labels and group probabilities for every sample $\*x \in \mathcal{D}_{\text{train}}$ for further training with G-DRO and PG-DRO respectively. In case of G-DRO, $\hat {\mathbb{Q}}(\*x)$ has all its mass in the predicted group. In Figure~\ref{fig:toy_img}, we visualize the decision boundary plot of models trained using G-DRO vs. PG-DRO. The vertical axis in the plots indicate the environmental/spurious feature ($z_{\text{e}}$) that highly correlates with $y$, and the horizontal axis is the invariant feature ($z_{\text{inv}}$). We observe that the PG-DRO trained model (Figure~\ref{fig:pgdro}) learns a more robust decision boundary (more dependent on invariant features) as compared to the model trained using G-DRO (Figure~\ref{fig:gdro}). Moreover, the improved worst-group test accuracy of using PG-DRO further validates the advantages of using group probabilities.

\begin{figure}[t]
\centering
\begin{subfigure}{0.5\columnwidth}
  \centering
  \includegraphics[scale = 0.25]{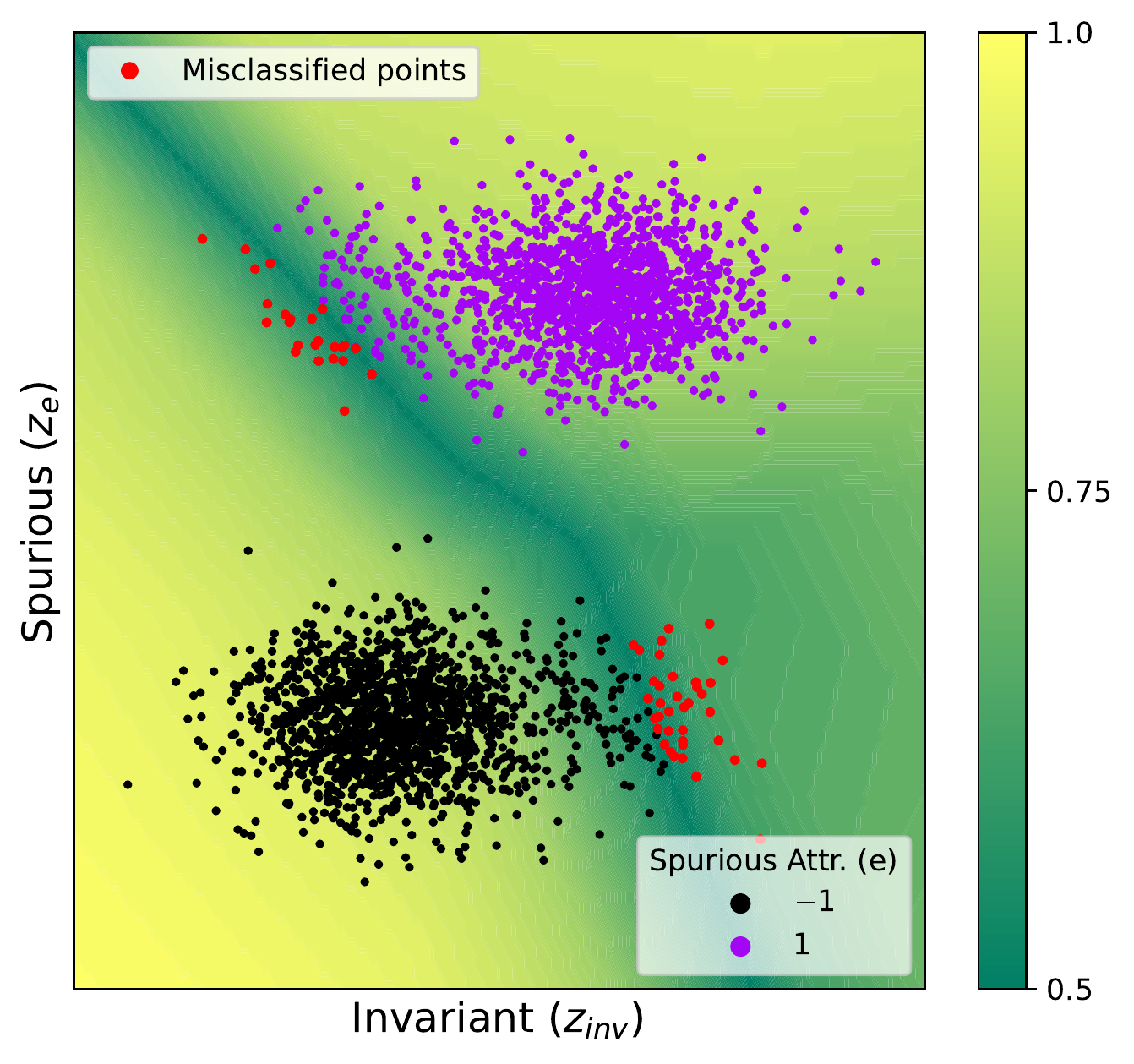}
\end{subfigure}%
\begin{subfigure}{0.5\columnwidth}
  \includegraphics[scale = 0.25]{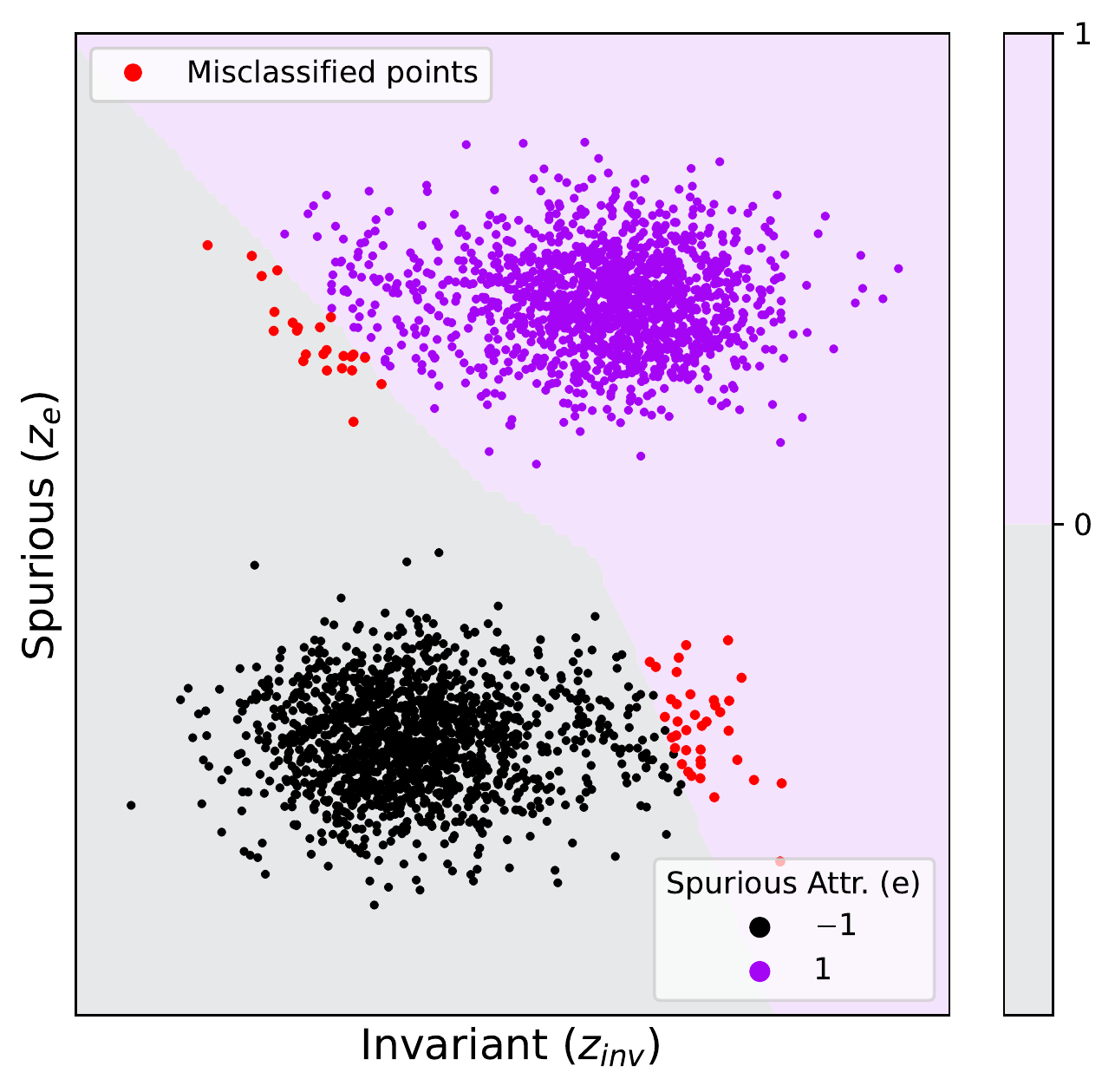}
\end{subfigure}
\caption{\small Consider data points $\*x \in \mathcal{D}_{\text{train}}$. Color of points encodes the true spurious attribute $e$. (Left) Confidence plot of the trained spurious attribute classifier. We visualize the maximum confidence ($\max_i \hat{\mathbb{P}}(e_i|x)$) of the spurious predictor. (Right) Decision boundary based on hard labeling. Color of background shade indicates the model's predicted environment ($\argmax_i \hat{\mathbb{P}}(e_i|x)$). Mis-classified points are marked in {red}, which are erroneously penalized with hard labeling. Using probabilistic group better captures uncertainty in spurious attribute prediction. }
\label{fig:mistake_img}
\end{figure}

\noindent \textbf{PG-DRO better captures uncertainty in spurious attribute predictor.} In Figure~\ref{fig:mistake_img}, we contrast the spurious attribute's prediction under soft (left) vs. hard (right) encoding. We can observe that the  classifier $\hat f_\text{env}$ can be uncertain on samples belonging to the minority groups, where spurious correlations do not hold. Hard labeling, in this scenario, does not consider the ambiguity in the predictions of the group classifier. As a result, the model can disproportionally penalize the samples, leading to poor worst-group robustness (See Figure~\ref{fig:gdro}). Using probabilistic group membership is more advantageous since it allows capturing group ambiguity and uncertainty in the spurious predictor. 

%% file: section/related_works.tex
\section{Related Works}
\label{sec: related_works}

\paragraph{Distributionally robust optimization.} A line of works~\cite{ben2013robust, wiesemann2014distributionally, blanchet2019quantifying, lam2015quantifying,namkoong2017variance, mohajerin2018data, bertsimas2018data} proposed algorithms to minimize the worst loss within a ball centered around the empirical distribution over training data. Recent works~\cite{sagawa2019distributionally, zhang2022correct} have raised concerns regarding these algorithms, as they tend to optimize for both the worst-group and average accuracy during training. Specifically, Sagawa \etal~\cite{sagawa2019distributionally} have shown that in the regime of over-parameterized models, DRO performs no better than empirical risk minimization.

\vspace{0.1cm}
\noindent\textbf{Improving robustness with group annotations.} When the group annotations are available for training data, one can leverage the information to alleviate the model's reliance on spurious correlations. Among this line of research, ~\citet{sagawa2019distributionally}, ~\citet{zhang2020coping}, ~\citet{mohri2019agnostic} have proposed designing objective functions to minimize the worst group training loss. In this study, we specifically constrain our attention to Group-DRO (G-DRO), an online optimization algorithm proposed by \citet{sagawa2019distributionally}, that focuses on training updates over data points from higher-loss groups. ~\citet{goel2020model} proposed CAMEL, which trains a CycleGAN~\cite{CycleGAN} model to learn data transformations and then apply it to extend the minority group through data augmentation synthetically. A group of methods~\cite{shimodaira2000improving, byrd2019effect, sagawa2020investigation} aims to improve the worst group performance through re-weighting or re-sampling the given dataset to balance the distribution of each group. Although these methods show promising results and improvement in worst-group performance over standard empirical risk minimization (ERM), there is a major caveat: all these methods assume the availability of group annotations for samples in the entire dataset---which can be expensive and prohibitive in practice.

\vspace{0.1cm}
\noindent \textbf{Improving robustness without group annotations.} Several recent works tackle the problem of reducing the dependency on group annotations for the entire training dataset. In particular, \citet{nam2020learning} proposed to train a pair of models simultaneously (``biased'' and ``debiased'' versions), such that their relative cross-entropy losses on each training example determine their importance weights in the overall training objective. ~\citet{liu2021just} first train an ERM model for a fixed number of epochs to identify samples being misclassified, and then train another model by up-weighting the misclassified samples. Similarly, EIIL~\cite{creager2021environment} and SSA~\cite{nam2022spread} first train a model to infer the group labels, and then use the generated group labels for robust training with G-DRO~\cite{sagawa2019distributionally}. Recently, a contrastive learning based method~\cite{zhang2022correct} was proposed for improving spurious correlations.
Another line of work includes re-weighting training samples based on weights determined through meta-learning~\cite{ren2018learning} or by learning explicitly using a small amount of group labeled samples~\cite{shu2019meta}. Our proposed approach is more related to methods belonging to this branch of study (which do not require group labels for the training set). However, our framework explores a novel optimization using the probabilistic group, instead of assigning a hard group label.

%% file: section/conclusion.tex
\section{Conclusion}
\label{sec:conclusion}

In this paper, we propose PG-DRO, a novel robust learning framework targeting the spurious correlation problem. Our framework explores the idea of probabilistic group membership for distributionally robust optimization. We broadly evaluate PG-DRO on both  computer  vision and NLP benchmarks, establishing superior performance among methods not using training set group annotations. With minimal group annotations, PG-DRO can favorably match and even outperform G-DRO (using group annotations for both training and validation set).
We hope our work will inspire future research to explore using probabilistic group for robust optimization.

%% file: section/acknowledge.tex
\section*{Acknowledgement}

We would like to thank Yifei Ming, Ziyang (Jack) Cai and Gabriel Gozum for insightful discussions and comments. We gratefully acknowledge
the support of the AFOSR Young Investigator Award under No. FA9550-23-1-0184; Philanthropic Fund from SFF; Wisconsin Alumni Research Foundation; faculty research awards from Google, Meta, and Amazon. Any opinions, findings, conclusions, or recommendations
expressed in this material are those of the authors and do not necessarily reflect the views, policies, or
endorsements either expressed or implied, of the sponsors.

%% file: section/CLIP.tex
\section{Zero-shot Group Labeling }
\label{sec:clip}
\begin{figure}[t]
\centering
\includegraphics[width=\columnwidth]{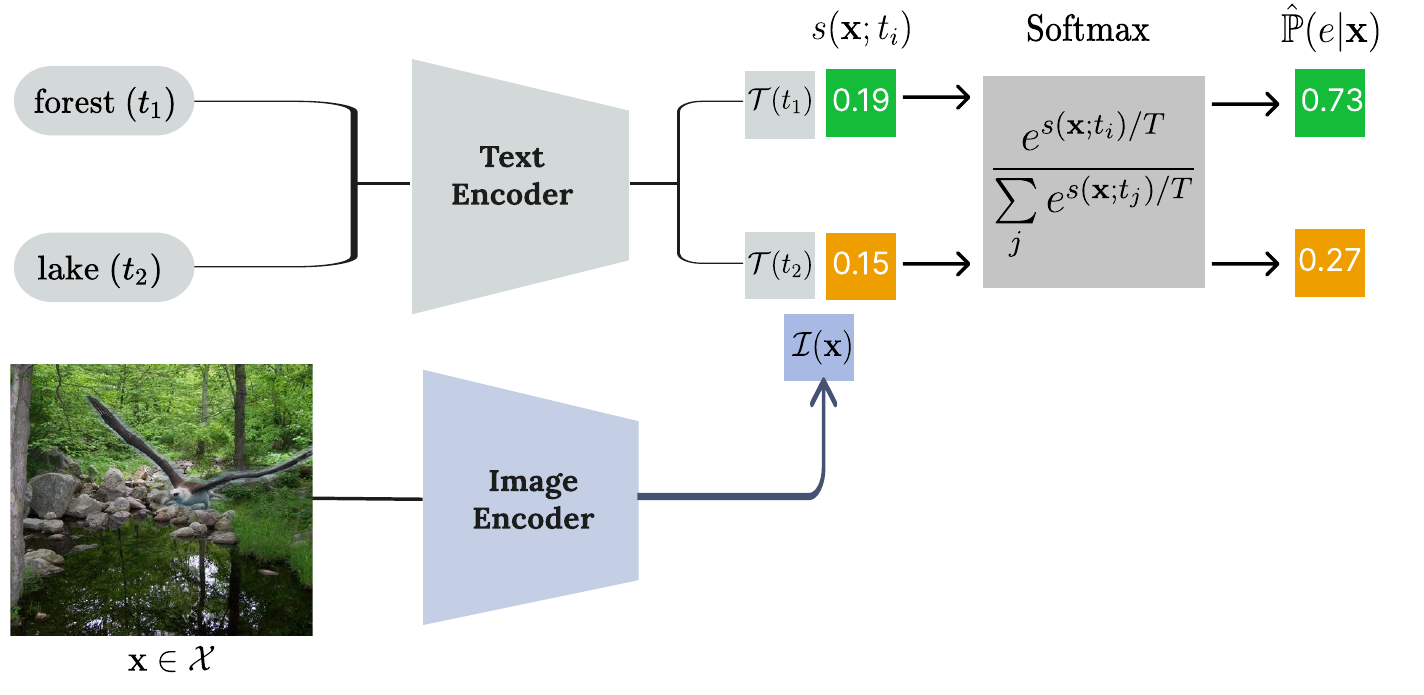}
\caption{\small Illustration of zeros-shot group labeling using CLIP~\cite{radford2021learning}. The input image $\*x$ is sampled from Waterbirds~\cite{sagawa2019distributionally}. We calculate cosine similarity, $s(\*x; t_i)$ between each text prompt $t_i$ and input image $\*x$ in the embedding space. Finally, we calculate $\hat{\mathbb{P}}(e|\*x)$ by normalizing the cosine similarities with a softmax function. }

\label{fig:proposed_approach_clip}
\end{figure}

In this section, we explore a new alternative for obtaining pseudo group probabilities by leveraging large-scale pre-trained models such as CLIP~\cite{radford2021learning}. Compared to the pseudo labeling approach in Section 3.2, here we provide a \emph{group-annotation-free} approach that is compatible with our PG-DRO learning objective. We briefly explain its working mechanism.
\paragraph{Zero-shot group labeling with CLIP.} 
As illustrated in Figure~\ref{fig:proposed_approach_clip}, we generate probabilistic group labels for each sample in a zero-shot fashion. We show that the CLIP-based approach can effectively alleviate the reliance on the availability of group labels. For reader's context, CLIP~\cite{radford2021learning} is trained on a dataset of 400 million text-image pairs. The training aligns an image with its corresponding textual description in the feature space. The vision-language representations have demonstrated superior performance on zero-shot image classification tasks. However, \emph{no prior work has explored its efficacy for pseudo group labeling.}

Our key idea is to match an input image to a set of environments $\mathcal{E}$. The pre-trained CLIP model consists of a text encoder $\mathcal{T}: t \rightarrow \mathbb{R}^d$ (e.g. Transformer~\cite{vaswani2017attention}) and an image encoder $\mathcal{I}: \*x \rightarrow \mathbb{R}^d$ (e.g. ViT~\cite{dosovitskiy2020image}).  Formally, given an input $\*x \in \X$ sampled from a set of environments $\mathcal{E} = \{e_1, e_2, \cdots, e_E\}$, we instantiate a set of text prompts $\{t_1, t_2, \cdots,t_E\}$, where $t_i$ represents textual description of environment $e_i$ in $\mathcal{E}$. For example, in the case of CelebA dataset, we have the set of text descriptions $\{\textsc{male}, \textsc{female}\}$.
The CLIP model uses cosine similarity  to measure the compatibility between the query image and each text prompt $t_i$: $s(\*x; t_i) = \frac{\mathcal{I}(\*x) \cdot \mathcal{T}(t_i)}{\lVert \mathcal{I}(\*x)\rVert \cdot \lVert \mathcal{T}(t_i) \rVert}$. Formally, the probabilistic group can be obtained by normalizing the cosine similarity with a softmax function:
\begin{equation}
  \hat{\mathbb{P}}(e_i |\*x) = \frac{e^{s(\*x; t_i)/T}}{\sum_{j=1}^E e^{s(\*x;t_j)/T}} .
\end{equation}

We then use $\hat{\mathbb{P}}(e|\*x)$ to define soft group probabilities $\hat{\mathbb{Q}}(\*x)$ and train with the PG-DRO objective, as described in Section 3. Note that our zero-shot approach only needs the meta information of environment names, which are more easily obtainable than sample-wise group annotations.  
\input{tables/clip_table}

\paragraph{Results. } In Table~\ref{tab:clip_comp}, we tabulate the performance of PG-DRO using zero-shot group labeling. We use PG-DRO (w/ CLIP) to differentiate approaches based on CLIP (as opposed to the default version in Section 3). 
We can draw two salient observations: \textbf{(1)} 
Zero-shot group labeling is fully compatible with our robust training objective and favorably matches the performance of G-DRO, which requires both training and validation sets. This is encouraging given there is no hand-annotated group labeling required in our approach. \textbf{(2)} PG-DRO (w/ CLIP) outperforms PG-DRO on both Waterbirds and CelebA datasets. Fundamentally, our method here differs from PG-DRO only in the pseudo-labeling technique. The improvement in worst-group test accuracy validates the efficacy of zero-shot group labeling.

%% file: tables/clip_table.tex
\begin{table*}[t]
\small
\centering
 \begin{tabular}{l ccccc}
\toprule
 \multirow{2}{2cm}{\textbf{Method}} &  \multirow{2}{*}{\centering Dataset with group label} & \multicolumn{2}{c}{\centering \textbf{Waterbirds}} & \multicolumn{2}{c}{\centering \textbf{CelebA}}\\ \cmidrule{3-6}
 & &  Avg. Acc & Worst Group Acc & Avg. Acc &  Worst Group Acc \\
 \midrule

 G-DRO~\cite{sagawa2019distributionally} & train \& val. set & 92.4$_{\pm{0.2}}$ & 90.7$_{\pm{0.9}}$ & 92.8$_{\pm{0.3}}$ & 88.7$_{\pm{1.5}}$\\
 \midrule
 PG-DRO (Ours)  & 5\% val. set & 91.7$_{\pm{0.2}}$ & 89.2$_{\pm{0.5}}$ & 92.0$_{\pm{0.3}}$ & 89.4$_{\pm{1.0}}$ \\
  \midrule
\scriptsize PG-DRO (w/ CLIP) (Ours) & \textbf{None} & 91.5$_{\pm{0.2}}$ & \textbf{89.4}$_{\pm{0.6}}$ & 92.7$_{\pm{0.2}}$ & \textbf{88.8}$_{\pm{0.9}}$ \\
 \bottomrule
 \end{tabular}
 \caption{\small Comparison of average and worst group test accuracies between methods leveraging CLIP model for pseudo labeling and those using supervised learning. Results (mean and std) of our proposed methods are estimated over 3 random seeds. Best performing results (among methods not using training set group annotations) are marked in \textbf{bold}.}
\label{tab:clip_comp}
\end{table*}

%% file: tables/pseudo_hyper.tex
\begin{table*}[h!]
\small
\centering
 \begin{tabular}{l c c c c c}
\toprule
 Dataset & Optimizer & Learning Rate ($lr$) & $l_2$ penalty & Epochs & Batch Size \\
 \midrule
 Waterbirds~\cite{sagawa2019distributionally} & SGD &  $10^{-3}$ & $10^{-4}$ & 200 & 64/16 \\
 CelebA~\cite{liu2015deep} & SGD & $10^{-4}$ & 0.1 & 300 & 64 \\
  MultiNLI~\cite{williamsnli} & ADAMW & $2\times10^{-5}$ & 0 & 50 & 32 \\
  CivilComments~\cite{borkancivil,koh2021wilds} & ADAMW & $10^{-5}$ & 0.01 & 50 & 8\\
 \bottomrule
 \end{tabular}
 \caption{\small Hyperparameter configurations used during pseudo labeling. To distinguish between configurations used when assuming group labeled annotations for whole validation set and $5\%$ of validation set, we use `val. set / 5\% val. set'.}
\label{tab:pseudo_hyper}
\end{table*}

%% file: tables/robust_hyper.tex
\begin{table*}[h!]
\small
\centering
 \begin{tabular}{l c c c c c c}
\toprule
 Dataset & Optimizer & Learning Rate ($lr$) & $l_2$ penalty & Epochs & Batch Size & $C$ \\
 \midrule
 Waterbirds~\cite{sagawa2019distributionally} & SGD &  $10^{-5}$ & 1 & 300 & 128 & 2/4 \\
 CelebA~\cite{liu2015deep} & SGD & $10^{-5}$ & 0.1 & 50 & 128 & 1/3\\
  MultiNLI~\cite{williamsnli} & ADAMW & $2\times10^{-5}$ & 0 & 10 & 32 & 1 \\
  CivilComments~\cite{borkancivil,koh2021wilds} & ADAMW & $10^{-5}$ & 0.01 & 5 & 16 & 0\\
 \bottomrule
 \end{tabular}
 \caption{\small Detailed hyperparameter configurations used during robust training. To distinguish between configurations used when assuming group labeled annotations for whole validation set and $5\%$ of validation set, we use `val. set / 5\% val. set' }
\label{tab:robust_hyper}
\end{table*}

%% file: tables/spurious_predictor.tex
\begin{table*}[h!]
\small
\centering
 \begin{tabular}{l c p{1.4cm}>{\centering\arraybackslash}p{1.4cm}>{\centering\arraybackslash}p{1.4cm}>{\centering\arraybackslash}p{1.4cm}}
\toprule
 \multirow{2}{2cm}{\textbf{Method}} &  \multirow{2}{3cm}{\centering Dataset with group label} & \multicolumn{2}{c}{\centering \textbf{Land Bird}} & \multicolumn{2}{c}{\centering \textbf{Water Bird}}\\ \cmidrule{3-6}
 & & Land & Water & Land &  Water \\
 \midrule
Supervised & val. set & 95.1 & 96.7 & \textbf{92.8} & 97.1 \\
 
Supervised &  5\% val. set & 89.4 & 94.1 & \textbf{96.4} & 90.7\\
\midrule
 Semi-Supervised~\cite{nam2022spread} & val. set & 94.7 & 96.2 & \textbf{92.8} & 96.4\\
 
Semi-Supervised~\cite{nam2022spread} &  5\% val. set & 83.7 & 95.1 & \textbf{91.1} & 93.3\\

\midrule
Zero-shot(w/ CLIP) & None & 97.3 & 88.6 & \textbf{83.9} & 98.3 \\
 
 \bottomrule
 \end{tabular}
 \caption{Comparison of group-wise accuracy of different pseudo-labeling
technique, on the Waterbirds~\cite{sagawa2019distributionally} dataset. Accuracy for the most under-represented group is marked in \textbf{bold}.}
\label{tab:spurious_pred_waterbirds}
\end{table*}

%% file: tables/runtime.tex
\begin{table*}[h!]
\caption{Runtime for each pseudo labeling technique for considered datasets.}
\label{tab:runtime}
\small
\centering
 \begin{tabular}{l p{2.2cm}>{\centering\arraybackslash}p{2cm}>{\centering\arraybackslash}p{2cm}>{\centering\arraybackslash}p{2.5cm}}
\toprule
 \textbf{Method} &  Waterbirds & CelebA & MultiNLI & CivilComments-WILDS \\
 \midrule

Semi-Supervised~\cite{nam2022spread} & 7.3 hrs & 5.2 hrs & 12.4 hrs & 22.8 hrs \\
\midrule
CLIP (Ours) & 0.2 hr & 0.6 hr & - & - \\
 
 \bottomrule
 \end{tabular}
\end{table*}

%% file: tables/analysis_vision.tex
\begin{table*}[h!]
\caption{\small Comparison of average and worst-group test accuracies for combinations of different group labeling approach and robust training methods when evaluated on image classification datasets: Waterbirds~\cite{sagawa2019distributionally} \& CelebA~\cite{liu2015deep}. Results (mean and std) of our proposed methods are estimated over 3 random seeds for consistency. }
\label{tab:analysis_vision_table}
\small
\centering
 \begin{tabular}{l c p{1.4cm}>{\centering\arraybackslash}p{1.5cm}>{\centering\arraybackslash}p{1.4cm}  >{\centering\arraybackslash}p{1.5cm}}
\toprule
 \multirow{2}{2cm}{\textbf{Method}} &  \multirow{2}{3cm}{\centering Dataset with group label} & \multicolumn{2}{c}{\centering \textbf{Waterbirds}} & \multicolumn{2}{c}{\centering \textbf{CelebA}}\\ \cmidrule{3-6}
 & &  Avg. Acc. & Worst Group Acc. & Avg. Acc. &  Worst Group Acc. \\
 \midrule
ERM & None & 97.3 & 63.2 & 95.6 & 47.2\\
\midrule
G-DRO~\cite{sagawa2019distributionally} & train \& val. set & 92.4$_{\pm{0.2}}$  & 90.7$_{\pm{0.9}}$ & 92.8$_{\pm{0.3}}$ & 88.7$_{\pm{1.5}}$\\
\midrule
SSA & val. set & 92.2$_{\pm{0.9}}$ & 89.0$_{\pm{0.6}}$ & 92.8$_{\pm{0.1}}$ & \textbf{89.8}$_{\pm{1.3}}$ \\
PG-DRO (Ours) & val. set & 92.4$_{\pm{0.4}}$ & \textbf{91.0}$_{\pm{0.8}}$ & 92.5$_{\pm{0.3}}$ & \textbf{89.8}$_{\pm{1.1}}$ \\
\midrule
 SSA & $5\%$ of val. set & 92.6$_{\pm{0.1}}$ & 87.1$_{\pm{0.7}}$ & 92.8$_{\pm{0.3}}$ & 86.7$_{\pm{1.1}}$ \\
 PG-DRO (Ours) & $5\%$ of val. set & 91.9$_{\pm{0.2}}$ & \textbf{89.2}$_{\pm{0.6}}$ & 91.7$_{\pm{0.3}}$ & \textbf{87.8}$_{\pm{1.0}}$ \\

\bottomrule

\end{tabular}
\end{table*}